\documentclass[lettersize,journal]{IEEEtran}
\usepackage{amsmath,amsfonts}
\usepackage{algorithmic}
\usepackage{algorithm}
\usepackage{array}
\usepackage[caption=false,font=normalsize,labelfont=sf,textfont=sf]{subfig}
\usepackage{textcomp}
\usepackage{stfloats}
\usepackage{url}
\usepackage{verbatim}
\usepackage{graphicx}
\usepackage{enumitem}
\usepackage{cite}
\usepackage{amssymb}
\usepackage{booktabs}
\usepackage{subcaption} 
\usepackage{xcolor}
\usepackage{adjustbox}
\usepackage{tabularx}
\usepackage{booktabs}
\usepackage{multirow}

\newcommand{\Lapl}{\mathbf{\mathop{\mathcal{L}}}}

\newcommand{\Mat}[1]{\boldsymbol{#1}}
\newcommand{\Set}[1]{\mathcal{#1}}

\newcommand{\wrt}{\emph{w.r.t. }}

\begin{document}

\title{Adaptive Self-supervised Robust Clustering for Unstructured Data with Unknown Cluster Number}

\author{Chen-Lu Ding\textsuperscript{$*$},
        Jiancan Wu\textsuperscript{$*$},
        Wei Lin,
        Shiyang Shen,
        Xiang Wang\textsuperscript{$\dagger$},
        Yancheng Yuan\textsuperscript{$\dagger$}
\thanks{$*$: Chen-Lu Ding and Jiancan Wu contribute equally to this manuscript.}
\thanks{$\dagger$: Xiang Wang and Yancheng Yuan are Corresponding Authors.}
\thanks{Chen-Lu Ding, Jiancan Wu, Wei Lin, and Xiang Wang are with University of Science and Technology of China. E-mail: dingchenlu200103@gmail.com, wujcan@gmail.com, kkwml@mail.ustc.edu.cn, xiangwang@ustc.edu.cn.}
\thanks{Shiyang Shen and Yancheng Yuan are with The Hong Kong Polytechnic University. E-mail: 22049485g@connect.polyu.hk, yancheng.yuan@polyu.edu.hk.}
}

\markboth{Journal of \LaTeX\ Class Files,~Vol.~0000, No.~0000}%
{Shell \MakeLowercase{\textit{et al.}}: A Sample Article Using IEEEtran.cls for IEEE Journals}

\maketitle

\begin{abstract}
We introduce a novel self-supervised deep clustering approach tailored for unstructured data without requiring prior knowledge of the number of clusters, termed Adaptive Self-supervised Robust Clustering (ASRC). In particular, ASRC adaptively learns the graph structure and edge weights to capture both local and global structural information. The obtained graph enables us to learn clustering-friendly feature representations by an enhanced graph auto-encoder with contrastive learning technique. It further leverages the clustering results adaptively obtained by robust continuous clustering (RCC) to generate prototypes for negative sampling, which can further contribute to promoting consistency among positive pairs and enlarging the gap between positive and negative samples. ASRC obtains the final clustering results by applying RCC to the learned feature representations with their
consistent graph structure and edge weights. Extensive experiments conducted on seven benchmark datasets demonstrate the efficacy of ASRC, demonstrating its superior performance over other popular clustering models. Notably, ASRC even outperforms methods that rely on prior knowledge of the number of clusters, highlighting its effectiveness in addressing the challenges of clustering unstructured data.
\end{abstract}

\begin{IEEEkeywords}
Clustering, unsupervised learning, self-supervised learning, convex clustering, adaptive graph structure learning.
\end{IEEEkeywords}

\section{Introduction}
\IEEEPARstart{W}{ith} the rapid development of the Internet, there has been an emergence of vast volumes of data. However, a significant portion of this data is unstructured, with the majority being not explicitly labeled. Clustering emerges as a pivotal technique in unsupervised learning.
By grouping similar data samples into coherent clusters and highlighting the distinctions among different clusters, it plays a crucial role in uncovering the intrinsic structures embedded within data. Classic clustering methods include center-based techniques (e.g., kmeans) \cite{lloyd1982least,arthur2007k,liu2010novel,hamalainen2020improving,celebi2013comparative}, graph Laplacian-based spectral clustering \cite{ng2001spectral,von2008consistency,lei2015consistency,von2007tutorial}, density-based clustering algorithms \cite{rodriguez2014clustering,campello2013density,ester1996density}, and hierarchical clustering methods \cite{johnson1967hierarchical,cheng2019hierarchical,rokach2005clustering}. 


However, these methods exhibit sensitivity to the initialization and suffer from the suboptimality of the obtained solutions due to the non-convex nature of their models. More critically, most of the popular clustering algorithms presuppose a prior knowledge or estimation of the cluster number, which proves impractical in real-world applications \cite{liu2023reinforcement}. Indeed, determining the optimal number of clusters presents a challenge as daunting as the clustering task itself.
To overcome the aforementioned challenges, the convex clustering model \cite{pelckmans2005convex,hocking2011clusterpath,lindsten2011clustering} has been recently proposed and has achieved notable success. 
This model aims to learn an approximate centroid for each data point via solving a strongly convex problem, where the fusion regularization terms encourage the merging of these centroids. It partitions the given data points into distinct clusters based on the obtained centroids, where the number of clusters is adjustable and a clustering path will be generated through the manipulation of the penalty parameter, with higher values generally yielding fewer clusters. 
In contrast to methods such as kmeans, where varying initial guesses of the cluster number may lead to unrelated outcomes, the solution to the convex clustering model is proved to be a continuous function of the penalty parameter \cite{chi2015splitting}. Recently, Shah and Koltun proposed a robust continuous clustering (RCC) model \cite{shah2017robust} that builds upon the convex clustering framework by incorporating an additional penalty function to the fusion terms. This penalty is designed to sever spurious connections between clusters in the constructed graph derived from the input data, thereby enhancing the model’s robustness.

Although the convex clustering model and the RCC model achieved some significant progress in addressing clustering tasks with unknown class numbers, they still suffer from some limitations that are interconnected:
\begin{itemize}
    \item
    \textbf{Poor-quality Feature Representations.}
    The simple yet elegant objective function of these models naturally requires informative feature representations of the data, which are often absent in practice. To address this, recent advancements in deep continuous clustering use auto-encoders to enhance feature learning \cite{shah2018deep}. Despite improvements, these models typically struggle to capture high-order structural information of the data, underscoring the need for learning clustering-friendly feature representations.

    \item 
    \textbf{Noisy Graph Structures.}
    The efficacy of these models is inherently tied to the estimated graph structure derived from unstructured data.
    To be more concrete, the constructed graph will guide the merging of centroids, as the fusion regularization terms penalize distances between the centroids of sample pairs within the same connected components of the graph. While the convex clustering model with a well-chosen graph and weights can recover true nonconvex clusters of data \cite{sun2021convex}, such as two half moon, it is known that the convex clustering model with a fully connected graph and uniform weights can only recover convex clusters \cite{nguyen2021convex} (which means that the convex hull of different clusters should be disjoint). This is strong evidence to demonstrate the importance of the graph structure. However, the graph structure, when generalized across diverse applications, often lacks a clear definition.
    Prior approaches typically build a k-nearest neighbors (kNN) graph based on the distance between raw features\cite{shah2017robust}, failing to adequately reflect global and high-order potential connectivities \cite{bo2020structural}. Such unclear signals may be amplified by the neighborhood aggregation scheme in GCNs, making the clustering procedure vulnerable to potential noise.

    \item
    \textbf{Inaccurate Regularization Weights.}
    The calibration of regularization weights matters in releasing the power of the convex clustering and RCC models \cite{chi2015splitting,nguyen2021convex,sun2021convex,dunlap2022local}, which is yet under-explored.
    Intuitively, the regularization weights among samples should be inversely proportional to their mutual distance, promoting consistency among cluster representatives for proximate samples \cite{sun2021convex}. Current practices often employ a Gaussian kernel function to assign weights based on feature similarity. While this can yield competitive clustering outcomes, its exponential form makes it highly sensitive to data scale, data noise, and spurious connections in the graph. 
\end{itemize}

In this paper, we introduce a new self-supervised deep clustering algorithm for unstructured data, termed Adaptive Self-supervised Robust Clustering (ASRC).
ASRC inherits the merits of the convex clustering and RCC models --- without the prerequisite of the cluster number, and addresses the aforementioned challenges in a unified manner.
Instead of directly constructing a kNN graph based on the input feature representations, ASRC adaptively adjusts the graph structure in a generative manner, where the learned probability \wrt graph edges can be naturally adopted by the RCC model, encouraging the capture of both local and global structural data characteristics.
The equipped graph of the unstructured data enables us to adopt the graph neural network GNN-based contrastive learning framework to learn clustering-friendly representations.
Inspired by RCC's superior performance, we refine the contrastive loss through selecting only out-of-cluster negative samples guided by the running RCC outcomes, circumventing the inclusion of potential false negatives. Finally, ASRC obtains the clustering results by applying RCC to the learned feature representations with their consistent graph structure and weights. Experimental results on multiple benchmark datasets demonstrate ASRC's superior performance, outperforming methods requiring prior cluster number knowledge. Our method's effectiveness and stability are enhanced through better feature representation, improved graph structure, and optimized weight assignment, as verified by ablation experiments.

Our main contributions can be summarized as follows,

\begin{itemize}
\item We propose a new adaptive self-supervised robust clustering method for unstructured data without requiring prior knowledge of the cluster number, which harnesses the advantages of RCC and addresses its limitations.

\item Our approach can learn clustering favorable feature representations by leveraging both structural information and contrastive signal. Particularly, we adopt RCC to adaptively generate prototypes for negative sampling, which enhances the performance of contrastive learning. 

\item Extensive experiments on benchmark datasets show that the proposed ASRC outperforms those requiring cluster numbers, with comprehensive ablation studies confirming its effectiveness. 
\end{itemize}

The rest of the paper is organized as follows. We introduce some notation and preliminaries of convex clustering and RCC models in Section \ref{sec: preliminary}. Details of our proposed ASRC model will be presented in Section \ref{sec: ASRC}. We present the numerical results in Section \ref{sec: numerical}. Detailed discussion of some related work can be found in Section \ref{sec: related-work} and we conclude the paper in Section \ref{sec: conclusion}.

\section{Preliminary}
\label{sec: preliminary}
\subsection{Notation}
Throughout this paper, we use boldface uppercase letters to denote matrices, e.g., $\boldsymbol{X}$; boldface lowercase letters represent vectors, e.g., $\boldsymbol{p}$.
A graph is represented as $\mathcal{G}=(\mathcal{V}, \mathcal{E}, \boldsymbol{X})$, where $\mathcal{V}, \mathcal{E}, \boldsymbol{X}$ are node set, edge set, and feature matrix, respectively. For each node $v_i\in\mathcal{V}$, its feature is represented by a $d$-dimensional vector $\boldsymbol{x}_i \in \mathbb{R}^d$.
$\boldsymbol{A} \in\mathbb{R}^{{n} \times {n}}$ is the adjacency matrix of the undirected weighted graph, whose elements represent the weights of the edges.
We denote $\widetilde{\boldsymbol{A}}=\boldsymbol{A}+\boldsymbol{I}$ the adjacency matrix with self-loop and $\widetilde{\boldsymbol{D}}$ $(\widetilde{\boldsymbol{D}}_{i i}=\sum_{j=1}^n \widetilde{\boldsymbol{A}}_{i j})$ the corresponding degree matrix, where $n$ denotes the number of nodes (or data points).
Let $|\cdot|$ denote the size of some set, $\|\cdot\|_2$ denote the $\ell_2$-norm (respectively, the spectral norm) of the vector (respectively, the matrix).

\subsection{Convex Clustering}
Given a collection of $n$ data points with feature matrix $\boldsymbol{X}=\left[\boldsymbol{x}_1, \boldsymbol{x}_2, \ldots, \boldsymbol{x}_n\right]^{\top} \in \mathbb{R}^{n \times d}$, the convex clustering model solves the following strongly convex problem:
\begin{equation}
\label{model: ccm}
\min_{\boldsymbol{U} \in \mathbb{R}^{n \times d}} ~ \frac{1}{2} \sum_{i=1}^n\left\|\boldsymbol{x}_i-\boldsymbol{u}_i\right\|_2^2 + \gamma\sum_{(i, j) \in \mathcal{E}} w_{ij} \left\|\boldsymbol{u}_{{i}}-\boldsymbol{u}_j\right\|_2,
\end{equation}
where $w_{ij} = w_{ji}$ are given weights, $\mathcal{E}$ is some given edge set of the constructed graph, and $\gamma \geq 0$ is the model hyperparameter. 

Due to the strong convexity of the objective function, the solution $\boldsymbol{U}^*(\gamma)$ is unique for any given $\gamma > 0$, which implies that the objective \eqref{model: ccm} is not sensitive to the initialization. It is apparent that $\boldsymbol{u}_i^*(0) = \boldsymbol{x}_i$, resulting in $n$ clusters if all the input data points are distinct. A larger value of $\gamma$ will push some columns of $\boldsymbol{U}_i^*(\gamma)$ to merge together, thereby reducing the cluster count. This observation motivates us to obtain a clustering path by sequentially solving the convex clustering objective \eqref{model: ccm} for a series of $\gamma$ values.


\subsection{Robust Continuous Clustering}
Drawing inspiration from convex clustering \cite{pelckmans2005convex}, RCC \cite{shah2017robust} employs robust estimators to optimize clustering representatives for each sample, without prior knowledge of the cluster number. Specifically, it solves the following optimization problem:
 
\begin{equation}
\label{rcc1}
\min_{\boldsymbol{U}\in \mathbb{R}^{n \times d}} ~ \frac{1}{2} \sum_{i=1}^n\left\|\boldsymbol{x}_i-\boldsymbol{u}_i\right\|_2^2+\frac{\lambda_1}{2} \sum_{(i, j) \in \mathcal{E}} w_{ij} \sigma\left(\left\|\boldsymbol{u}_{{i}}-\boldsymbol{u}_j\right\|_2\right) ,
\end{equation}
where $\sigma(\cdot)$ is the penalty function and $\lambda_1$ is the trade-off parameter balancing
the regularization term. The Geman-McClure estimator $\sigma(x)=\frac{\alpha x^2}{\alpha+x^2}$ \cite{geman1987statistical} is a default choice for the RCC model \cite{shah2017robust}, where $\alpha$ is the scalar parameter. In this setting, it follows from \cite{shah2017robust} that we can then solve the following equivalent optimization problem (\wrt $\boldsymbol{U}$): 
\begin{equation}
\begin{aligned}
\label{rcc2}
&\min_{\boldsymbol{U}, \boldsymbol{L}} ~ \mathcal{O}(\boldsymbol{U},\boldsymbol{L})=  \frac{1}{2} \sum_{i=1}^n\left\|\boldsymbol{u}_i-\boldsymbol{x}_i\right\|_2^2 \\
& + \frac{\lambda_1}{2} \sum_{(i, j) \in \mathcal{E}} w_{ij}\left(l_{i j}\left\|\boldsymbol{u}_{\boldsymbol{i}}-\boldsymbol{u}_{\boldsymbol{j}}\right\|_2^2+\alpha\left(\sqrt{l_{i j}}-1\right)^2\right)
,
\end{aligned}
\end{equation}
where $L = (l_{i j})_{(i, j) \in \mathcal{E}}$ is the auxiliary variable. It is worth mentioning that the equivalent problem \eqref{rcc2} has several advantages. On the one hand, it is clear from this formula that the trade-off between the penalty terms $l_{ij}\left\|\boldsymbol{u}_{\boldsymbol{i}}-\boldsymbol{u}_{\boldsymbol{j}}\right\|_2^2$ and $\alpha\left(\sqrt{l_{i j}}-1\right)^2$ in \eqref{rcc2} can disregard some spurious links by forcing some $l_{i j} = 0$, where the meaning of the parameter $\alpha$ becomes clear. On the other hand, it can naturally apply an alternating minimization algorithm to solve \eqref{rcc2}, where each sub-problem can be solved efficiently. In particular, when we fix $\boldsymbol{U} = \widetilde{\boldsymbol{U}}$ and minimize $\mathcal{O}(\widetilde{\boldsymbol{U}},\boldsymbol{L})$ \wrt $\boldsymbol{L}$, the closed-form solution is given as
\begin{equation}
\label{l_solution}
l_{i j}=\left(\frac{\alpha}{\alpha+\left\|\tilde{\boldsymbol{u}}_i-\tilde{\boldsymbol{u}}_{j}\right\|_2^2}\right)^2 .
\end{equation}
Meanwhile, we can minimize $\mathcal{O}(\boldsymbol{U}, \widetilde{\boldsymbol{L}})$ \wrt $\boldsymbol{U}$ for a fixed $\widetilde{\boldsymbol{L}}$ by solving the linear system 
\begin{equation}
\label{uproblem}
\boldsymbol{S}\boldsymbol{U}=\boldsymbol{X} ,
\end{equation}
where \begin{equation}
\label{balance}
\boldsymbol{S}=\boldsymbol{I}+\lambda_1 \sum_{(i, j) \in \mathcal{E}} w_{ij} \tilde{l}_{i j}\left(\boldsymbol{e}_{\boldsymbol{i}}-\boldsymbol{e}_{\boldsymbol{j}}\right)\left(\boldsymbol{e}_{\boldsymbol{i}}-\boldsymbol{e}_{\boldsymbol{j}}\right)^{\top} .
\end{equation}
Here, $\boldsymbol{e}_i \in \mathbb{R}^n$ is the $i$-th column of the identity matrix $\boldsymbol{I} \in \mathbb{R}^{n \times n}$. 

It is worth noting that, in the RCC model \cite{shah2017robust}, the regularization parameter $\lambda_1$ is adaptively updated as 
\begin{equation}
\label{lambda}
\lambda_1=\frac{\|X\|_2}{\left\|\sum_{(i, j) \in \mathcal{E}} w_{ij} l_{i j}\left(\boldsymbol{e}_{\boldsymbol{i}}-\boldsymbol{e}_{\boldsymbol{j}}\right)\left(\boldsymbol{e}_{\boldsymbol{i}}-\boldsymbol{e}_{\boldsymbol{j}}\right)^{\top}\right\|_2}.
\end{equation}
Due to its dynamic update mechanism, RCC eliminates the need for manually adjusting the penalty parameter. More details can be found in \cite{shah2017robust}.

\section{Adaptive Self-supervised Robust Clustering}
\label{sec: ASRC}
\begin{figure*}[htbp]
\centering
\includegraphics[width=0.78\textwidth]{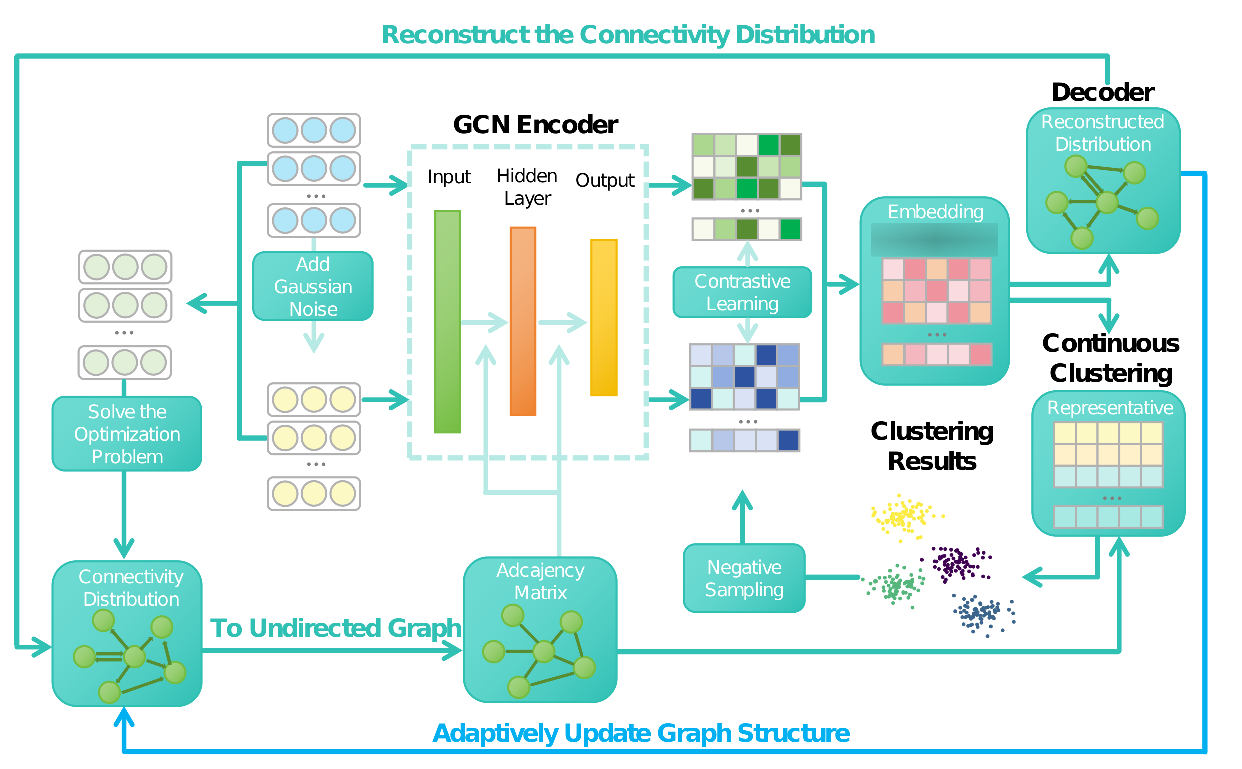}
\caption{The framework of ASRC. Firstly, we augment features to obtain multiple views of data and address problem \eqref{pij} to generate the weighted graph. Then, we apply GAE to obtain embeddings. The graph structure is updated adaptively by increasing the sparsity parameter $k$. After updating the new graph structure, we incorporate contrastive learning to retrain GAE. Once the training process converges, we utilize the learned graph structure, edge weights, and representations as inputs for robust continuous clustering, ultimately obtaining clustering results. Then we use the clustering results to guide the selection of negative samples and update the clustering results until convergence.}
\label{framework}
\end{figure*}

Our proposed ASRC model consists of three key modules: (a) enhanced adaptive graph structure learning module, (b) self-supervised feature representation learning module, and (c) enhanced robust continuous clustering module. An overview paradigm of ASRC is shown in Fig. \ref{framework}.
We adaptively learn a graph structure for the unstructured data guided by the principle that the probability of a connection between two nodes in the graph should be inversely proportional to the distance between the current embeddings of the two nodes. The equipped graph then enables us to design a GNN-based contrastive learning framework to learn clustering favorable feature representations. Importantly, we generate the prototypes by RCC to guide the negative sampling in contrastive learning which can enhance its performance. The RCC model applied to the learned feature representations with their consistent graph structure and compatible weights can thus yield superior clustering results. It is worthwhile emphasizing that prior knowledge of the number of underlying clusters is not required in ASRC. Our approach inherits the advantages of RCC yet addresses its limitations.

\subsection{Enhanced Adaptive Graph Structure Learning}

The core of deep clustering methods lies in learning a clustering favorable representation. Conceptually, the clustering task begins by forming local cluster structures, and these sub-clusters are subsequently merged to form larger clusters, a process that relies on high-order structural information. This parallels the neighborhood aggregation pattern in graph neural networks, which can uncover high-level structural information while also paying attention to local features. Therefore, leveraging graph neural networks to learn discriminative representations is meaningful.

However, unstructured data often lacks clear graph structures. The commonly used graph constructed using kNN usually fails to capture global topological information. To address this challenge, we adopt an enhanced adaptive graph structure learning scheme, which can adaptively learn the weighted graph structure of the unstructured data. The learned graphs also enable us to adaptively refine the feature representations using a GNN-based auto-encoder. We call our designed framework an enhanced adaptive graph neural network based auto-encoder (EadaGAE). Importantly, compared to adaGAE proposed in \cite{li2021adaptive}, the weighted graph learned by EadaGAE can be naturally adopted in the RCC model to yield a clustering result. Next, we will describe the details of EadaGAE.

\subsubsection{Weights Updating}
As discussed earlier, the kNN graph simply generates the graph based on the distances between raw features, without considering global and high-order potential connectivity information. The unclear signals will be amplified after the neighbor aggregation operation of GCN, leading to unsatisfactory embeddings \cite{wu2021self}. Instead, EadaGAE, which inherits the advantages of adaGAE, will adaptively and consistently update the graph structure and the embeddings of the nodes, thereby addressing this issue. 

Specifically, adaGAE interprets the weights of edge $(i, j)$ in the directed graph as the conditional probability $p(v_j~|~v_i)$, or $\Mat{p}_{ij}$ for brevity. Therefore, a weighted directed graph can be constructed by learning the conditional probabilities. For convenience, we denote $\boldsymbol{p}_i=[  p\left(v_1 \mid v_i\right), p\left(v_2 \mid v_i\right), \ldots, p\left(v_n \mid v_i\right)]$ as the probabilities that node $v_i$ is connected to other nodes. Note that $\boldsymbol{p}_i \in \Delta_n$, where $\Delta_n$ is the simplex defined as $\Delta_n := \{\boldsymbol{x} \in \mathbb{R}^n ~|~ \boldsymbol{x} \geq 0$ and $\sum_{j=1}^n \boldsymbol{x}_{j} = 1\}$. 


According to the homogeneity theory of graphs, connected nodes exhibit similar structural characteristics. That is, if the value of $\Mat{p}_{ij}$ is larger, then the representations of $v_i$ and $v_j$ should be more similar. Therefore, the problem of graph construction can be transformed into the following optimization problem \cite{li2021adaptive}:

\begin{equation}
\label{pij}
\min_{\boldsymbol{p}_{i} \in \Delta_n} \sum_{i=1}^n\left\langle\boldsymbol{p}_{i}, \boldsymbol{d}_{i}\right\rangle+\frac{\gamma_i}{2}\left\|\boldsymbol{p}_{i} - \boldsymbol{q}_{i}\right\|_2^2.
\end{equation}
Here, a regularization term is added to avoid trivial solutions.
Each element in $\Mat{d}_i$, or $\boldsymbol{d}_{i j} = \|\boldsymbol{z}_i - \boldsymbol{z}_j\|_2$, is the distance between the representations of $v_i$ and $v_j$, $\Mat{z}_i$ denotes the representation of $v_i$.
$\{\gamma_i\}_{i=1}^{n}$ and $\{ \Mat{q}_i \}_{i=1}^{n}$ are the set of trade-off parameters and prior distributions, respectively. Following the settings of adaGAE \cite{li2021adaptive}, we adopt the uniform distribution as the prior distribution in objective \eqref{pij}. In supplementary to \cite{li2021adaptive}, we explain two reasons for adopting the prior distribution $\boldsymbol{q}_i$ as the uniform distribution. 

On the one hand, to ensure that points within the same group are closely connected by edges, we prefer to learn a sparse graph to better capture the structures of the data. The adoption of $\boldsymbol{q}_i$ as uniform distribution gives an explicit meaning of the regularization parameter $\gamma_i$ in terms of the sparsity level of the solution to \eqref{pij}. In particular, we can guarantee the solution $\bar{\boldsymbol{p}}_i$ ($1 \leq i \leq n$) has no more than $k$ non-zero entries by setting 
\begin{equation}
\label{eq: val-gamma}
\gamma_i=\frac{1}{2} \sum_{m=1}^k\left(\boldsymbol{d}_{i .}^{(k+1)} - \boldsymbol{d}_{i .}^{(m)}\right),
\end{equation}
where $\boldsymbol{d}_{i .}^{(m)}$ ($1 \leq m \leq n$) represents the $m$-th smallest item in $\boldsymbol{d}_i$. Moreover, under this scenario, it follows from \cite{nie2014clustering} that the solution to \eqref{pij} is given by
\begin{equation}
\label{graph_solution}
\boldsymbol{p}_{ij}^* =\frac{\left(\boldsymbol{d}_{i .}^{(k+1)}-\boldsymbol{d}_{i j}\right)_{+}}{k \boldsymbol{d}_{i .}^{(k+1)}-\sum_{m=1}^k \boldsymbol{d}_{i .}^{(m)}},
\end{equation}
where $(\cdot)_{+}$ denotes the Relu function.  Therefore, the graph structure can be updated through Eq.(\ref{graph_solution}) efficiently. We will adaptively update the sparsity level $k$ to refine the graph structure, which we will discuss in detail later.

On the other hand, we prefer that the solution to the optimization problem \eqref{pij} should keep the homogeneity of the graph. In other words, we require $\boldsymbol{p}^{*}_{ij} \geq \boldsymbol{p}^{*}_{il}$ if the embeddings satisfying $\boldsymbol{d}_{ij} \leq \boldsymbol{d}_{il}$. Indeed, setting $\boldsymbol{q}_i$ as uniform distribution is an appropriate choice to satisfy this requirement. In order to prove this, we need to derive a solution to \eqref{pij} for general prior distributions $\boldsymbol{q}_i$. We include the details in Appendix D.

\subsubsection{Representation Learning} Given the current embeddings $\boldsymbol{X}$ and the sparsity level $k$, we can utilize Eq.(\ref{graph_solution}) to update the edge weights of the directed graph. We refer to this graph as the raw graph, with $\boldsymbol{P}$ representing the corresponding edge weight matrix. For simplicity, we convert the raw graph into an undirected graph, where $\boldsymbol{A}=\left(\boldsymbol{P}^{\top}+\boldsymbol{P}\right) / 2$. Unlike traditional GNNs that directly add self-loops, the self-connections in our graph are adaptively formed based on the sparsity of each node. Thus, $\widetilde{\boldsymbol{A}}$ can be directly set to $\boldsymbol{A}$.


We initially employ a multi-layer GCN as the encoder to extract high-order information. Taking a $2$-layer GCN as example, the resultant representation matrix $\Mat{Z}$ is defined as,
\begin{equation}
\label{gcn}
\boldsymbol{Z}=\boldsymbol{\hat{A}}\left( \varphi(\boldsymbol{\hat{A}} \boldsymbol{X}\boldsymbol{\Theta^{(1)}})\right)\boldsymbol{\Theta^{(2)}} ,
\end{equation} 
where $\boldsymbol{\hat{A}}=\boldsymbol{\tilde{D}}^{-\frac{1}{2}} \boldsymbol{\tilde{A}} \boldsymbol{\tilde{D}}^{-\frac{1}{2}}$ is the normalized graph matrix and $\varphi(\cdot)$ is the activation function. $\boldsymbol{\Theta}$ is the trainable parameter of GCN.

Most decoders aim to reconstruct the adjacency matrix through the inner product. However, this formulation disregards the distance between nodes and fails to effectively extract structural information from the representation. Here we reconstruct the connectivity distribution between nodes based on the Euclidean distance among representations. The reconstructed connectivity distribution is as follows,
\begin{equation}
\label{softmax}
\hat{\boldsymbol{p}}_{ij} =\frac{e^{-\|\boldsymbol{z}_i - \boldsymbol{z}_j\|_2}}{\sum_{l=1}^n e^{-\|\boldsymbol{z}_i - \boldsymbol{z}_l\|_2}} ,
\end{equation}
where $\boldsymbol{z}_i$ is the embedding of $v_i$ obtained by Eq.\eqref{gcn}. Therefore, it is a natural idea to adopt the Kullback-Leibler divergence to measure the disparity between the two distributions. Combining the two objectives of consistent representation and reconstructing the connectivity probability distribution, our final graph-based optimization problem can be written as

\begin{equation}
\label{gaefinal}
\min_{\boldsymbol{\Theta}} ~ \mathcal{L}_{GAE}(\boldsymbol{\Theta})=\sum_{i=1}^n \sum_{j=1}^n \boldsymbol{p}_{i j} \log \frac{\boldsymbol{p}_{i j}}{\hat{\boldsymbol{p}}_{i j}}+\frac{\lambda_2}{2}  \sum_{i=1}^n \sum_{j=1}^n \boldsymbol{p}_{i j} \boldsymbol{d}_{i j},
\end{equation}
where $\lambda_2$ is a hyper-parameter to control the trade-off. Note that when optimizing the parameters of GAE, the graph structure remains fixed. Therefore, the regularization term in objective \eqref{pij} can be dropped. 

In adaGAE\footnote{The details of adaGAE can be found in Algorithm \ref{algada}.} \cite{li2021adaptive}, node embeddings are updated by solving the optimization problem \eqref{gaefinal}, and the graph structure is refined by increasing the value of $k$. This strategy stems from the observations reported in \cite{li2021adaptive} that when $k$ is small, better reconstruction is often achieved at the cost of poor updates in embeddings.
To avoid this collapse, adaGAE iteratively increases $k$, ensuring that samples within the same cluster are connected by edges as much as possible, thereby capturing high-level information.
Initially, $k$ is set as $k=k_0$, and it is linearly updated as
\begin{equation}
\label{k_update}
k = k + s
\end{equation}
after solving the optimization problem \eqref{gaefinal}, with a preset maximum update number $T_1$. The values of $k_0$, $s$, and $T_1$ will be discussed in the numerical experiments. 

However, direct adoption of adaGAE \cite{li2021adaptive} is not appropriate for our purpose of using RCC to finally yield the clustering results.
Instead, for a given value of $k$, we propose to iteratively refine the embeddings by optimizing \eqref{gaefinal} and update the graph structure by Eq.\eqref{graph_solution} until converged. Subsequently, we will dynamically adjust the value of $k$ as needed until the training of both the graph structure and GAE is completed. Such enhancement not only guarantees the convergence of the algorithm but also maintains alignment between the graph structure and the learned representation.
More importantly, unlike the adaGAE algorithm shown in lines $4-10$ of Algorithm \ref{algada}, we conduct an additional update of the graph weights using Eq.\eqref{graph_solution} with the newly optimized embeddings and the current $k$ value, which can be found in lines $12-13$ of {{Algorithm \ref{alg}}}.
This step, though not required in other clustering methods such as kmeans adopted in \cite{li2021adaptive}, is crucial for the RCC model in \eqref{rcc1}, as it ensures the graph structure, weights, and representations remain consistent, thereby preventing incorrect clustering results. We term our enhanced adaptive graph learning module EadaGAE in short. 


\begin{algorithm}[htbp]
\caption{Algorithm to optimize adaGAE.}
\begin{algorithmic}[1]
\label{algada}
\STATE {\textbf{Input:}} Date $\boldsymbol{X}$; initial sparsity $k_0$; increment of sparsity $s$; the number of iterations $T_1$ for updating graph structure with different $k$ values.

\STATE{\textbf{Output:}} Clustering results.

\STATE{\textbf{Initialization:}} $k = k_0$; $\boldsymbol{z}_i = \boldsymbol{x}_i$.

\STATE{\textbf{While}} current iteration not reach $T_1$ \textbf{do}
\STATE \hspace{0.5cm} Compute $d\left(v_i, v_j\right)$ and update raw weight matrix $\boldsymbol{P}$
\STATE \hspace{0.5cm} with Eq.(\ref{graph_solution}).

\STATE\hspace{0.5cm} {\textbf{Repeat}}
\STATE \hspace{1.0cm} Update the weights of GAE with Eq.(\ref{optimize}) by 
\STATE \hspace{1.0cm} gradient descent.

\STATE\hspace{0.5cm} {\textbf{Until}} convergence or reach the maximum iterations.
\STATE\hspace{0.5cm} Get new embedding $\boldsymbol{Z}$.
\STATE\hspace{0.5cm} Update the sparsity with Eq.(\ref{k_update}).

\STATE{\textbf{End}} 
\STATE{Run kmeans on $\boldsymbol{Z}$ and get the clustering result.}

\end{algorithmic}
\label{alg1}
\end{algorithm}

\subsection{Self-supervised Learning}
In EadaGAE, we strategically increase the value of $k$ to maximize the connectivity within clusters, thus avoiding collapse. However, it may make node embeddings between different clusters less discriminative. This may be harmful to the performance of the convex clustering model and RCC model, since the ratio of the maximum intra-cluster distance and the minimum inter-centroid distance plays a crucial role in successfully recovering the underlying clusters \cite{sun2021convex,panahi2017clustering}.
To address this limitation, self-supervised learning emerges as a promising enhancement. The core principle of self-supervised learning --- to attract similar instances closer while distancing dissimilar ones --- naturally aligns with the objectives of clustering.
Therefore, we integrate self-supervised learning, particularly through contrastive learning, to achieve a more discriminative feature representation.

Specifically, contrastive learning is characterized by two key components: (1) augmenting data instances, which generates multiple views of each instance, and (2) optimizing contrastive loss, which maximizes the agreement between different views of the same instance, compared to that of dissimilar instances.
Following this, we first create an augmented view $\Mat{X}_2$ of each original feature $\Mat{X}_1$\footnote{$\Mat{X}_1 = \Mat{X}$; subscript added for clarity.} through the addition of Gaussian noise.
Each pair of features is then encoded to produce respective representations via Equation~\eqref{gaefinal}, yielding $\Mat{Z}_1$ and $\Mat{Z}_2$.
These two representation views are fused in a linear manner as $\Mat{Z} = \frac{1}{2}\left( \Mat{Z}_1 + \Mat{Z}_2 \right)$ to form the final representation, which is utilized to compute the reconstruction loss~\eqref{gcn} throughout the training process.
Note that at each training steps, we are aware of the ongoing clustering results. This feedback provides valuable signal for alleviating the sampling bias \cite{zhao2021graph}, which has been shown to impair the effectiveness of contrastive learning.
Intuitively, instances residing in the same cluster tend to be potential false negatives.
Accordingly, we refine the commonly used InfoNCE loss \cite{liu2023hard} with a clustering-guided adjustment, formalized as follows:
\begin{equation}
    \Lapl_{ssl} = -\frac{1}{2n}\sum_{i=1}^{2n} \log \frac{e^{s\left(\Mat{z}_i^{\prime}, \Mat{z}_i^{\prime\prime}\right)}}{e^{s\left(\Mat{z}_i^{\prime}, \Mat{z}_i^{\prime\prime}\right)} + \sum\limits_{j \in \Set{N}_{\setminus i}} \left( e^{s\left(\Mat{z}_i^{\prime},\Mat{z}_j^{\prime}\right)}+e^{s\left(\Mat{z}_i^{\prime},\Mat{z}_j^{\prime\prime}\right)}\right)},
\end{equation}
where $\left(\Mat{z}_i^{\prime}, \Mat{z}_i^{\prime\prime}\right) \in \{(\Mat{z}_i^{(1)}, \Mat{z}_i^{(2)}), (\Mat{z}_i^{(2)}, \Mat{z}_i^{(1)}) \}$, $\Mat{z}_i^{(1)}$ and $\Mat{z}_i^{(2)}$ are the $i$-th row of $\Mat{Z}_1$ and $\Mat{Z}_2$, respectively.
The similarity function $s(\cdot)$, such as cosine similarity in this work, evaluates the resemblance between features, and $\Set{N}_{\setminus i}$ excludes instances from the same cluster as sample $i$ in current step.

This type of contrastive learning aids in acquiring more resilient and discriminative feature representations. Additionally, as previously discussed, enhanced representations can facilitate the learning of graph structures and lead to improved weights, thereby significantly enhancing clustering performance.

\subsection{Adaptive Self-supervised Robust Clustering}
As mentioned earlier, in order to generate better graph structures for obtaining discriminative representations, we first adapt the graph structure to align with clustering objectives and utilize GAE to extract high-level information. This tailored graph structure then serve as the scope (edge set $\mathcal{E}$) of regularization in Problem \eqref{rcc2}, with the transformed average weights $\widetilde{\boldsymbol{P}}$ acting as the regularization weights, further matching the clustering task. Finally, we enhance the representation quality through a clustering-guided contrastive learning. It is notable that, since weighted graph learning and node representation learning are carried out alternately, the quality of the graph structure, weights, and representation mutually reinforce each other.
Overall, the objective of ASRC is formulated as,
\begin{equation}
\label{optimize}
\mathcal{L}_{ASRC}=\mathcal{L}_{GAE}+\beta \mathcal{L}_{ssl},
\end{equation}
where $\beta$ is the trade-off parameter. We employ gradient descent to optimize Eq.\eqref{optimize}, enabling the consistent improvement of the graph structure, weights, and representations.
Thus, problem \eqref{rcc2} can be rewritten as,

\begin{equation}
\begin{aligned}
\label{rcc3}
&\mathcal{O}(\boldsymbol{U},\boldsymbol{L})=  \frac{1}{2} \sum_{i=1}^n\left\|\boldsymbol{u}_i-\boldsymbol{z}_i\right\|_2^2 \\
& + \frac{\lambda_1}{2} \sum_{(i, j) \in \mathcal{E}} \widetilde{\boldsymbol{P}}_{ij}\left(l_{i j}\left\|\boldsymbol{u}_{\boldsymbol{i}}-\boldsymbol{u}_{\boldsymbol{j}}\right\|_2^2+\alpha\left(\sqrt{l_{i j}}-1\right)^2\right)
,
\end{aligned}
\end{equation}
where $\widetilde{\boldsymbol{P}}_{ij}$ denotes the element in the $i$-th row and $j$-th column of matrix $\widetilde{\boldsymbol{P}}$.

We should emphasize that the entire training process of ASRC is unsupervised and does not depend on any label information.
The detailed training procedure of ASRC is illustrated in Algorithm \ref{alg}.

\begin{algorithm}[htbp]
\caption{Algorithm to optimize ASRC.}\label{alg:alg1}
\begin{algorithmic}[1]
\label{alg}
\STATE {\textbf{Input:}} Date $\boldsymbol{X}$; initial sparsity $k_0$; increment of sparsity $s$; the number of iterations $T_1$ for updating graph structure with different $k$ values; the number of iterations $T_2$ for updating graph structure with fixed $k$ values; the number of iterations $T_3$ for clustering; iteration interval $t$ for updating $\lambda_1$ and $\alpha$; trade-off parameter $\lambda_2$ and $\beta$; learning rate of GAE $\eta$; clustering threshold $\delta$.

\STATE{\textbf{Output:}} Clustering results.

\STATE{\textbf{Initialization:}} $k = k_0$; $\boldsymbol{z}_i = \boldsymbol{x}_i$.

\STATE{\textbf{While}} current iteration not reach $T_1$ \textbf{do}
\STATE \hspace{0.4cm} Compute $d\left(v_i, v_j\right)$ and update raw weight matrix $\boldsymbol{P}$
\STATE \hspace{0.4cm} with Eq.(\ref{graph_solution}).

\STATE \hspace{0.4cm} \textbf{While} current iteration not reach $T_2$ \textbf{do}

\STATE\hspace{0.8cm} {\textbf{Repeat}}
\STATE \hspace{1.2cm} Update the weights of GAE with Eq.(\ref{optimize}) by 
\STATE \hspace{1.2cm} gradient descent.

\STATE\hspace{0.8cm} {\textbf{Until}} convergence or reach the maximum iterations.

\STATE \hspace{0.8cm} Compute $d\left(v_i, v_j\right)$ and update raw weight matrix 
\STATE \hspace{0.8cm} $\boldsymbol{P}$ with Eq.(\ref{graph_solution}).
\STATE \hspace{0.4cm} \textbf{End}
\STATE\hspace{0.4cm} Update the sparsity with Eq.(\ref{k_update}).

\STATE{\textbf{End}} 

\STATE{Obtain the raw edge weight matrix $\boldsymbol{P}$ and node representation $\boldsymbol{Z}$.}

\STATE{\textbf{Initialization:}} $\boldsymbol{u}_i = \boldsymbol{z}_i$; $l_{ij = 1}$; $\alpha \gg \max \left\|\boldsymbol{z}_i-\boldsymbol{z}_j\right\|_2^2$; initialize $\lambda_1$ via Eq.(\ref{lambda}).
\STATE{\textbf{While}} not convergence or current iteration not reach $T_3$ \textbf{do}
\STATE\hspace{0.2cm} Update $l_{ij}$ with Eq.(\ref{l_solution}) and update $\boldsymbol{S}$ with Eq.(\ref{balance}).
\STATE\hspace{0.2cm} Every $t$ iterations, update $\lambda_1$ with Eq.(\ref{lambda}) and 
\STATE\hspace{0.2cm} update $\alpha$ with $\alpha=\max \left(\frac{\alpha}{2}, \frac{\delta}{2}\right)$.

\STATE{\textbf{End}} 

\STATE Connect node pairs that meet $\left\|\boldsymbol{u}_i-\boldsymbol{u}_j\right\|_2<\delta$, and clusters can be given by the connectivity structure.

\end{algorithmic}
\label{alg1}
\end{algorithm}
\subsection{Computational Complexity}
The time complexity of the neighbor aggregation step $\boldsymbol{\hat{A}} \boldsymbol{X}\boldsymbol{\Theta^{(i)}} \in \mathbb{R}^{n \times d_i}$ in GCN training process is $O\left(l|\mathcal{E}| \sum d_i\right)$, where $l$ is the total number of gradient descent and $\mathcal{E}$ is the number of edges. The computational complexity for building the adjacency matrix of the graph $\boldsymbol{A} \in \mathbb{R}^{n \times n} $ is $O\left(n^2\right)$. For RCC, it has been demonstrated in \cite{shah2017robust} that its per-iteration time complexity is in linear order \wrt the number of samples $n$ and the input feature dimension, making it highly efficient.

\section{Experiment}
\label{sec: numerical}
\subsection{Experimental Settings}
\subsubsection{Datasets} In this paper, we conduct extensive experiments across 7 unstructured benchmark datasets, including texts, images, and protein expressions. The statistics of these datasets are summarized in Table \ref{dataset}. Here we provide a brief introduction to these datasets:
\textbf{20NEWS} is a widely used text dataset for text classification and text mining, containing 20 different news thematic categories. We adopt the configuration in \cite{li2021adaptive} and select the first four groups for experiments;
\textbf{UMIST} \cite{hou2013joint} comprises $564$ pictures of $20$ individuals, each represented through frontal and side views captured from multiple perspectives;
The Columbia Object Image Library (\textbf{COIL-20}) \cite{nene1996columbia} includes grayscale images of $20$ different objects, each imaged from various angles;
\textbf{MNIST} is a collection of handwritten digit images. Following the setting of \cite{li2021adaptive}, we only use MNIST-$test$, and to maintain notation simplicity, MNIST-$test$ is represented as MNIST.
\textbf{JAFFE} \cite{lyons1999automatic} contains $213$ images depicting the facial expressions of $10$ Japanese women.
\textbf{Mice Protein} \cite{asuncion2007uci} dataset measures the protein expression levels in the cerebral cortex of control and trisomic mice, spanning eight distinct groups. 
Note that the images from UMIST, COIL-$20$, and JAFFE are downsampled to $32\times32$ pixels. All features across these datasets are normalized to a range of $0$ to $1$. 

\begin{table}[htbp]
\caption[]{The statistics of the datasets.}
\label{dataset}
\renewcommand{\arraystretch}{1.2}
\begin{tabular}{@{\hspace{1.5em}}cl@{\hspace{1em}}cl@{\hspace{1.7em}}cl@{\hspace{1.5em}}c@{\hspace{1.5em}}}
\toprule
\textbf{Datasets}     &  & \# \textbf{Size} &  & \# \textbf{Features} &  & \# \textbf{Classes} \\ 
\midrule
\textbf{20NEWS}     &  & 2823    &  & 8000        &  & 4          \\
\textbf{UMIST} \cite{hou2013joint}       &  & 575     &  & 1024        &  & 20         \\
\textbf{COIL-20} \cite{nene1996columbia}     &  & 1440    &  & 1024        &  & 20         \\
\textbf{MNIST}        &  & 10000   &  & 784         &  & 10         \\
\textbf{JAFFE} \cite{lyons1999automatic}       &  & 213     &  & 1024        &  & 10         \\
\textbf{Mice Protein} \cite{asuncion2007uci} &  & 552     &  & 77          &  & 8         \\
\textbf{USPS}         &  & 1854    &  & 256         &  & 10         \\ 
\bottomrule
\end{tabular}
\end{table}

\subsubsection{Baseline}
We select seven state-of-the-art methods for comparison\footnote{Due to the absence of pre-trained weights, we exclude DCC \cite{shah2018deep} from our experiments.}, encompassing four categories.

\noindent \textbf{- Traditional center-based clustering methods}:
\begin{itemize}
    \item \textbf{Kmeans \cite{hartigan1979algorithm}}: It is a conventional center-based clustering method utilizing distance measurement.
\end{itemize}

\noindent\textbf{- Continuous clustering methods}:
\begin{itemize}
    \item \textbf{RCC \cite{shah2017robust}}: It introduces robust estimator, which is an enhanced version of convex clustering.
\end{itemize}

\noindent\textbf{- Graph spectral clustering methods}:
\begin{itemize}
    \item \textbf{N-Cut \cite{shi2000normalized}}: It maps data points to a low-dimensional feature space by solving the eigenvectors of the Laplacian matrix. 

    \item \textbf{SpectralNet \cite{shaham2018spectralnet}}: It employs deep learning to perform scalable and generalizable spectral clustering by embedding data into the eigenspace of a graph Laplacian.
\end{itemize}

\noindent\textbf{- GNN-based clustering methods}:
\begin{itemize}
    \item \textbf{GAE \cite{kipf2016variational}}: It combines GCN encoder and kmeans to obtain clustering results.

    \item \textbf{MaskGAE \cite{li2022maskgae}}: It utilizes masked graph modeling (MGM) to enhance representation learning.
    
    \item \textbf{SDCN \cite{bo2020structural}}: SDCN leverages a delivery operator to transfer representations from autoencoders to GCN layers. 
    
    \item \textbf{AdaGAE \cite{li2021adaptive}}: It adaptively constructs graphs to effectively exploit high-level information and non-Euclidean structures.
\end{itemize}

\subsubsection{Evaluation Metric}
To evaluate the clustering results, we adopt the following metrics: 
\begin{itemize}
\item \textbf{Adjusted Mutual Information (AMI)}: It is a metric commonly used in clustering analysis to quantify the agreement between clustering results and ground truth labels.
\item \textbf{Adjusted Rand Index (ARI)}: ARI not only considers agreement but also takes into account the effects of random assignment and cluster imbalance. 
\end{itemize}
For each metric, a higher value indicates a better clustering result.




\subsection{Implementation Details}
For fair comparisons, we reproduce the results either from the officially released code\footnote{We follow https://github.com/yhenon/pyrcc to reproduce RCC.} or reimplement the method with careful fine-tuning as suggested in the original paper.
For GNN-based methods, we employ kNN to construct the graph structure followed by the kmeans clustering on the trained embeddings to facilitate clustering. The criterion for constructing the KNN graph is chosen from \{euclidean, cosine\} depending on the dataset.
Except for RCC, all other algorithms require prior knowledge of the number of clusters.
Regarding N-Cuts \cite{shi2000normalized}, we generate the similarity matrix by applying Gaussian kernel (RBF) as follows,
\begin{equation}
s_{i j}=\frac{\exp \left(-\frac{\left\|\boldsymbol{x}_i-\boldsymbol{x}_j\right\|_2^2}{\sigma}\right)}{\sum_{j \in \mathcal{N}_i} \exp \left(-\frac{\left\|\boldsymbol{x}_i-\boldsymbol{x}_j\right\|_2^2}{\sigma}\right)}, 
\end{equation}
where $\mathcal{N}_i$ is the set of node $i$'s $k$-nearest neighbors and $\sigma$ is the Gaussian kernel weight.
Following \cite{li2021adaptive}, we set the number of layers of GCN to $2$ to prevent the over-smoothing issue.
The activation function is set to ReLU. For all methods, we conduct a grid search for the optimal learning rate within the set of \{$1e-4$, $1e-3$, $1e-2$\}. The parameter search ranges for each model are specified as follows:
For GAE, the hidden layers are adjusted between [$256$, $128$] and [$128$, $64$].
For SDCN, the number of epochs for pretraining and training are tuned in \{$10$, $20$, $30$, $40$, $50$, $60$, $70$, $80$, $90$, $100$\} and \{$100$, $200$, $300$\} respectively. The pretrain dimension is explored in \{$300$, $400$, $500$, $600$\}, and the feature vector dimension is adjusted in \{$10$, $20$, $30$, $40$, $50$\}.
For MaskGAE, the encoder and decoder dropout rates are tuned in \{$0.6$, $0.7$, $0.8$, $0.9$\} and \{$0.1$, $0.2$, $0.3$, $0.4$, $0.5$\} respectively. The balance parameter $\lambda$ in AdaGAE is tuned in \{$1$, $1e-1$, $1e-2$, $1e-3$\}. The exact parameter settings of ASRC are shown in the appendix B. To mitigate the influence of randomness, we report the averaged results in ten runs.
It should be noted that RCC \cite{shah2017robust} exhibits a high sensitive to distance metrics. Extensive preliminary empirical results suggest that the application of PCA significantly improves performance on the 20NEWS and COIL-$20$ datasets. To ensure fairness, we apply PCA uniformly on all methods for these two datasets. The analysis and experimental results of PCA components are shown in the appendix C.



\begin{table}[t]
    \centering
    \caption[]{The average number of clusters reported by RCC, ASRC for each dataset and the ground truth.}
    \label{number}
    \renewcommand{\arraystretch}{1.2}
    \begin{tabular}{cl@{\hspace{2.5em}}cl@{\hspace{2.5em}}cl@{\hspace{2.5em}}c}
    \toprule
    \textbf{Datasets}     &  \multicolumn{1}{c}{\textbf{\# Classes (RCC)}} &  \multicolumn{1}{c}{\textbf{\# Classes (ASRC)}}  & \multicolumn{1}{c}{\textbf{\# Classes}} \\ 
    \midrule
    \textbf{20NEWS}     & \multicolumn{1}{c}{2586}    & \multicolumn{1}{c}{5}         & \multicolumn{1}{c}{4}          \\
    \textbf{UMIST}         & \multicolumn{1}{c}{25}     & \multicolumn{1}{c}{21}        & \multicolumn{1}{c}{20}         \\
    \textbf{COIL-20}       & \multicolumn{1}{c}{54}     & \multicolumn{1}{c}{20}        & \multicolumn{1}{c}{20}         \\
    \textbf{MNIST}         & \multicolumn{1}{c}{13}    & \multicolumn{1}{c}{11.4}          & \multicolumn{1}{c}{10}         \\
    \textbf{JAFFE}          & \multicolumn{1}{c}{16}      & \multicolumn{1}{c}{11}        & \multicolumn{1}{c}{10}         \\
    \textbf{Mice Protein}  & \multicolumn{1}{c}{79}      & \multicolumn{1}{c}{18}           & \multicolumn{1}{c}{8}         \\
    \textbf{USPS}          & \multicolumn{1}{c}{33}      & \multicolumn{1}{c}{11.4}          & \multicolumn{1}{c}{10}         \\ 
    \bottomrule
    \end{tabular}
\end{table}

\begin{table*}[htbp]
\renewcommand{\arraystretch}{1.3}
\caption[]{Clustering results on seven datasets (mean±std). The best results in each case are marked in bold. The results are presented in percentage format.}
\label{main result}
\begin{tabular}{c|c|c|cc|cccc|cc}
\specialrule{1pt}{0pt}{0pt}

\multirow{2}{*}{Datasets} & \multirow{2}{*}{Metric} & \multicolumn{1}{c|}{Center-based}  & \multicolumn{2}{c|}{Spectral Clustering}  & \multicolumn{4}{c|}{GNN-based Clustering} & \multicolumn{2}{c}{Continuous Clustering}\\ \cline{3-11}

&     & Kmeans    &   N-Cut     & SpectralNet  &     SDCN   & GAE        & MaskGAE      & AdaGAE & RCC & ASRC             \\ \hline
\multirow{2}{*}{20NEWS}            & AMI    & 14.03±0.05 &  2.89±0.35 &  29.17±0.42  & 11.36±0.76 & 7.13±0.74 & 26.51±0.03   & 15.22±0.00 &   0.03±0.00  & \textbf{32.10±0.00}  \\
             & ARI    & 3.84±0.05  &  2.66±0.58  &  22.40±0.79  & 3.74±3.76 & 4.66±0.25 & 18.49±0.04   & 13.83±0.00 &  0.27±0.00   & \textbf{26.55±0.00}  \\ \hline
             
\multirow{2}{*}{UMIST}             & AMI    & 60.96±2.10 & 82.29±0.00 & 62.53±1.61 &   56.92±3.17 &68.49±2.64 &61.25±1.31
     & 90.26±0.01 & 79.47±0.00 & \textbf{94.42±0.00}  \\
             & ARI     & 33.13±2.39 & 64.53±0.00 & 35.20±1.28 &   29.84±2.80 &43.52±3.24 &36.43±1.41    & 79.52±0.02 & 52.32±0.00 & \textbf{89.65±0.01}  \\ \hline
             
\multirow{2}{*}{COIL-20}             & AMI    & 78.81±1.56 & 93.85±0.00 & 75.07±2.14 &   82.08±0.04 &82.32±1.31 &79.16±0.93 & 97.76±0.00 & 84.24±0.00 & \textbf{97.97±0.00}  \\
             & ARI     & 62.94±4.24 & 77.61±0.00 & 54.13±1.65 &   69.07±1.66 &68.46±3.24 &64.03±2.21 & 94.75±0.00 & 76.36±0.00 & \textbf{96.24±0.00}  \\ \hline
             
\multirow{2}{*}{MNIST}             & AMI    & 50.13±0.11 &71.59±0.00 & 71.97±0.32 &  69.48±0.60 &62.91±1.27 &63.90±0.24  & 81.90±0.01 & 78.93±0.00 & \textbf{82.54±0.00} \\
             & ARI    & 38.26±0.12 &45.83±0.00 & 60.73±0.46 &  60.20±1.56 &49.36±1.96 &53.60±0.38 & 74.59±0.03 & 60.54±0.00 & \textbf{77.19±0.01} \\ \hline
             
\multirow{2}{*}{JAFFE}             & AMI     & 90.51±1.27 &95.09±0.00 & 81.47±2.54 &   60.26±3.14 &94.00±1.91 &86.71±1.14 & 94.44±0.00 &  92.21±0.00 & \textbf{96.39±0.00}  \\
             & ARI     & 83.49±2.25 &92.27±0.00 & 66.47±0.98 &   34.34±3.38 &90.49±3.14 &76.69±1.76 & 91.50±0.00 &  87.70±0.00 & \textbf{95.57±0.00}  \\ \hline
             
\multirow{2}{*}{Mice Protein}             & AMI     & 35.33±1.13 &40.18±0.00 & 38.43±2.49 &  46.70±2.81 &49.64±2.88 &39.96±2.56
 & 55.84±0.01 &  58.24±0.00 & \textbf{70.20±0.02}  \\
                                          & ARI    & 19.23±0.37 &16.71±0.00 & 21.10±1.50 &  26.21±2.88 &28.58±2.13 &23.41±1.86 & 37.84±0.02 &  21.50±0.00 & \textbf{45.39±0.02}  \\ \hline
             
\multirow{2}{*}{USPS}            & AMI     & 62.40±0.40 &  74.78±0.00 & 71.50±0.52 &   78.38±1.65 &75.75±0.83 &70.79±2.77   & 82.94±0.01 &  79.55±0.00 & \textbf{84.12±0.00}  \\
             & ARI     & 53.25±1.48 & 57.36±0.00 & 56.40±0.62 &   73.01±4.42 &66.96±0.81 &61.44±4.40     & 76.33±0.01 &  73.21±0.00 & \textbf{79.56±0.00}           \\ \specialrule{1pt}{0pt}{0pt}
\end{tabular}
\end{table*}

\subsection{Comparison with Baselines}
 The clustering results of ASRC and all baselines on $7$ benchmark datasets are shown in Table \ref{main result}. Table \ref{number} reports the number of clusters obtained by RCC and ASRC on different datasets. From these results, we can obtain the following observations:

\begin{itemize}

\item Compared with other baselines, the `std' value of RCC \cite{shah2017robust} across all datasets consistently remains at $0$, demonstrating the stability of RCC. This can be attributed to RCC's design, which does not rely on specific initialization. ASRC, building upon the strengths of RCC, similarly exhibits stable clustering results without any extreme variations.


\item  Among GNN-based methods, adaptive graph structure learning approaches, such as adaGAE \cite{li2021adaptive} and ASRC, consistently outperform the traditional graph construction method like kNN.
The primary limitation of kNN lies in its overly simplistic nature, failing to capture the underlying structural information. It relies solely on low-level information, misguiding the information propagated within GNNs.
In contrast, the adaptive graph learning strategy excels in learning high-level structural information, thereby enhancing representation quality for clustering.

\item On the 20NEWS dataset, the AMI and ARI values of RCC are close to $0$, indicating poor performance. Given that 20NEWS is a text dataset, its features are inherently discrete, failing to sufficiently capture the relationships among samples.
The ambiguous graph structure and poor-quality representations further exacerbate this issue, resulting in RCC's tendency to produce clusters with numerous outliers and very few samples.
This can be confirmed from Table \ref{number} that the cluster number of RCC on 20NEWS is extremely large, almost $600$ times greater than the actual number of classes. 

\item Compared to RCC, ASRC provides more accurate predictions regarding the number of clusters. This suggests that ASRC is more adept at discerning the underlying clustering patterns within unstructured data.
The enhanced adaptive graph structure learning and clustering-guided negative sampling strategy contribute to the generation of representations that are more conducive to clustering.
Consequently, embeddings of the same clusters exhibit greater similarity, while those of different clusters become more distinguishable.

\item The proposed ASRC consistently outperforms all baselines across all cases, demonstrating its superiority. Specifically, compared to the strongest baseline, ASRC achieves a notable increase of 6.09\% in AMI, 9.12\% in ARI on average. Unlike GNN-based clustering methods that utilize kNN to construct graphs, enhanced adaptive graph structure learning can better reveal the edge connectivity information between samples. In contrast to adaGAE, the consistent graph structure and weight assignment boost the performance of continuous clustering. When compared to RCC, ASRC benefits from the enhanced feature representation, improved graph structure, and optimized weight assignment, resulting in clearer and more accurate clustering results.
\end{itemize}

\begin{table*}[h]
\renewcommand{\arraystretch}{1.3}
\caption[]{Clustering results with different variants on 20NEWS, Mice Protein, and UMIST (mean±std). The best results in each case are marked in bold. The results are presented in percentage format.}
\label{ablation}
\begin{center}
\begin{tabular}{@{\hspace{1.5em}}c@{\hspace{1em}}|@{\hspace{2em}}c@{\hspace{2em}}|@{\hspace{4em}}c@{\hspace{4em}}c@{\hspace{4em}}c@{\hspace{4em}}c@{\hspace{4em}}c@{\hspace{3em}}}
\specialrule{1pt}{0pt}{0pt}
Datasets  & Metric & RCC  & AdaGAE   &   ASRC-1 &  ASRC-2   & ASRC             \\ \hline
\multirow{2}{*}{20NEWS}             & AMI    & 0.03±0.00 &  15.22±0.00 &  29.32±0.01  &   30.06±0.01   &  \textbf{32.10±0.00} \\
             & ARI    & 0.27±0.00  &  13.83±0.00   &  23.83±0.02 &  24.41±0.02 & \textbf{26.55±0.00}\\ \hline
             
\multirow{2}{*}{UMIST}             & AMI    & 79.47±0.00 &  90.26±0.01  &   91.28±0.00  &   91.42±0.00 &  \textbf{94.42±0.00} \\
             & ARI    & 52.32±0.00 &  79.52±0.02  &   79.69±0.01  &   82.22±0.01 &  \textbf{89.65±0.01}\\ \hline
             
\multirow{2}{*}{Mice Protein}             & AMI    & 58.24±0.00 &  56.83±0.01  &  65.28±0.02  &   68.66±0.01 &  \textbf{70.20±0.02}\\
             & ARI    & 21.50±0.00 &  37.84±0.02  &  37.34±0.01  &   44.37±0.01 &  \textbf{45.39±0.02}\\ \hline
             
\specialrule{1pt}{0pt}{0pt}
\end{tabular}
\end{center}
\end{table*}

\subsection{Further Analysis}

\subsubsection{Ablation Studies}
The enhanced graph structure learning and debiased negative sampling are two key designs in ASRC, we here design the following ablation experiments to validate their effectiveness. Specifically, we incorporate two variants of ASRC: the variant \textbf{ASRC-1} excludes both the enhanced graph structure learning and debiased negative sampling, while the variant \textbf{ASRC-2} omits only the debiased negative sampling.
We also report the results of \textbf{RCC} and \textbf{adaGAE}, where RCC constructs the graph structure using kNN and computes weights using the Gaussian kernel function; adaGAE adaptively constructs graphs to effectively exploit high-level information. Due to space constraints, Table. \ref{ablation} only presents the experiment results on three datasets: 20NEWS, UMIST, and Mice Protein. The complete experimental results for all datasets can be found in the appendix A. We can observe that:


\begin{itemize}
\item Compared to RCC, ASRC-1, ASRC-2, and ASRC exhibit substantial performance improvements. This underscores the necessity of adaptive graph structure learning.
Unlike traditional kNN graphs, which often suffer from noise, adaptive graph structure learning facilitates the capture of both local and global structural data characteristics. The resultant weighted graph more accurately reflects the relationships between samples, making it more suitable for use as weights in the regularization terms of convex clustering.

\item  ASRC-2 outperforms ASRC-1 in most cases, demonstrating the superiority of enhanced graph structure. The enhancement to adaGAE ensures that the graph structure, weights, and representations are consistent. This consistency leads to more accurate and accordant inputs for RCC, thereby improving clustering results.

\item  ASRC consistently surpasses ASRC-2, suggesting the effectiveness of debiased negative sampling. Specifically, such clustering-guided adjustment mitigates sampling bias by addressing the issue where instances from the same cluster might be incorrectly identified as false negatives, resulting in more distinctive embeddings.
\end{itemize}

In conclusion, the integration of enhanced adaptive graph structure learning and debiased negative sampling enables ASRC to learn better graph structures, consistent weights, and discriminative data representations, thereby culminating in improved clustering results.

\subsubsection{Hyperparameter Sensitivity Analysis}
To examine the impact of various hyperparameters, we conduct experiments focusing on the increase in sparsity parameter $s$, the number of iterations for updating the graph structure $T_1$, and the trade-off parameters $\beta$ and $\lambda_2$. Due to space constraints, we provide the clustering results on the UMIST dataset exclusively in Fig. \ref{para}. We have the following observations:
\begin{figure}[htbp]
    \centering
    \begin{minipage}[b]{0.22\textwidth}
        \centering
        \includegraphics[width=\textwidth]{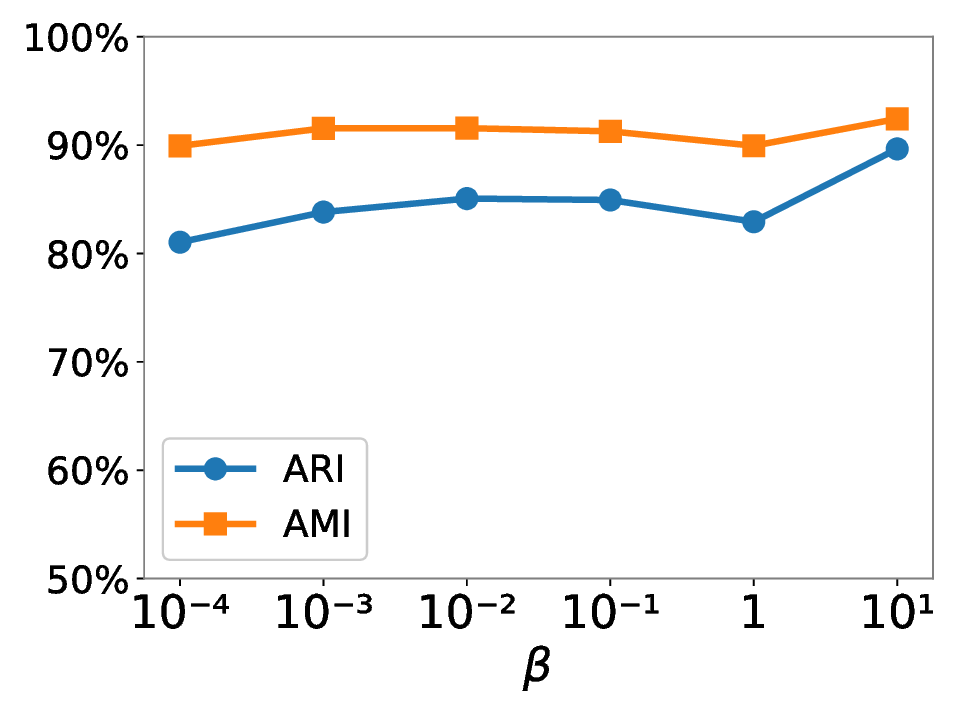}
    \end{minipage}
    \begin{minipage}[b]{0.22\textwidth}
        \centering
        \includegraphics[width=\textwidth]{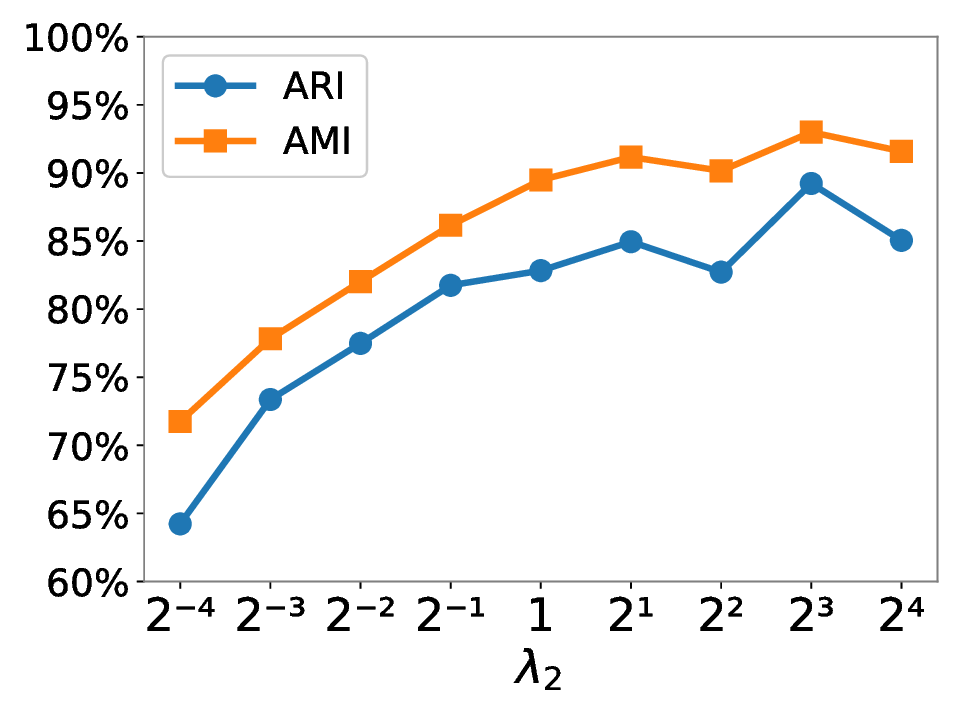}
    \end{minipage}   
    \begin{minipage}[b]{0.22\textwidth}
        \centering
        \includegraphics[width=\textwidth]{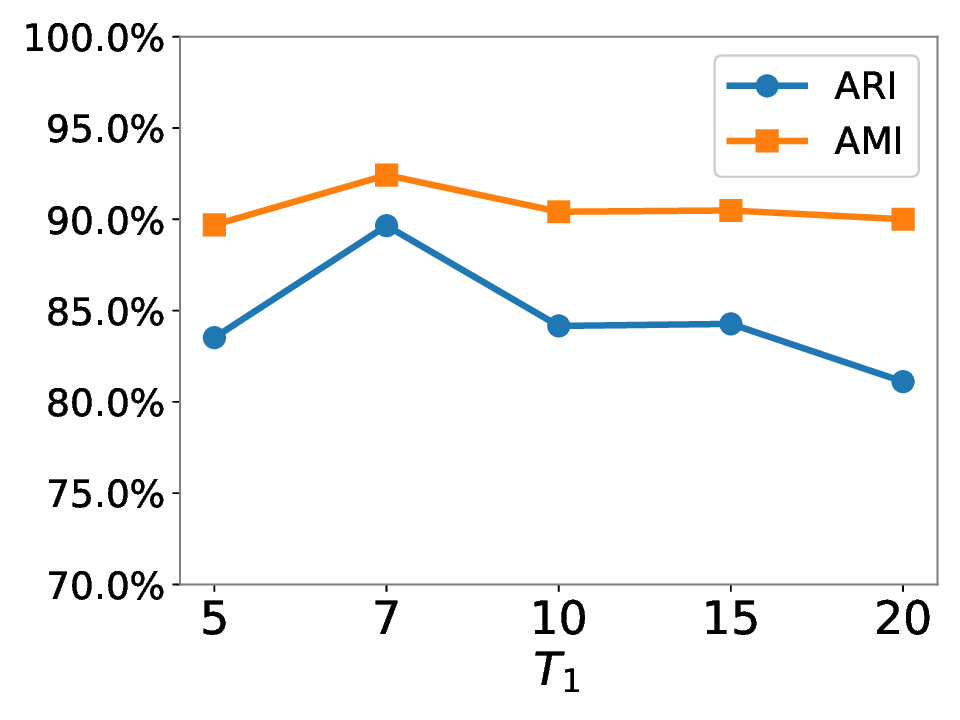}
    \end{minipage}
    \begin{minipage}[b]{0.22\textwidth}
        \centering
        \includegraphics[width=\textwidth]{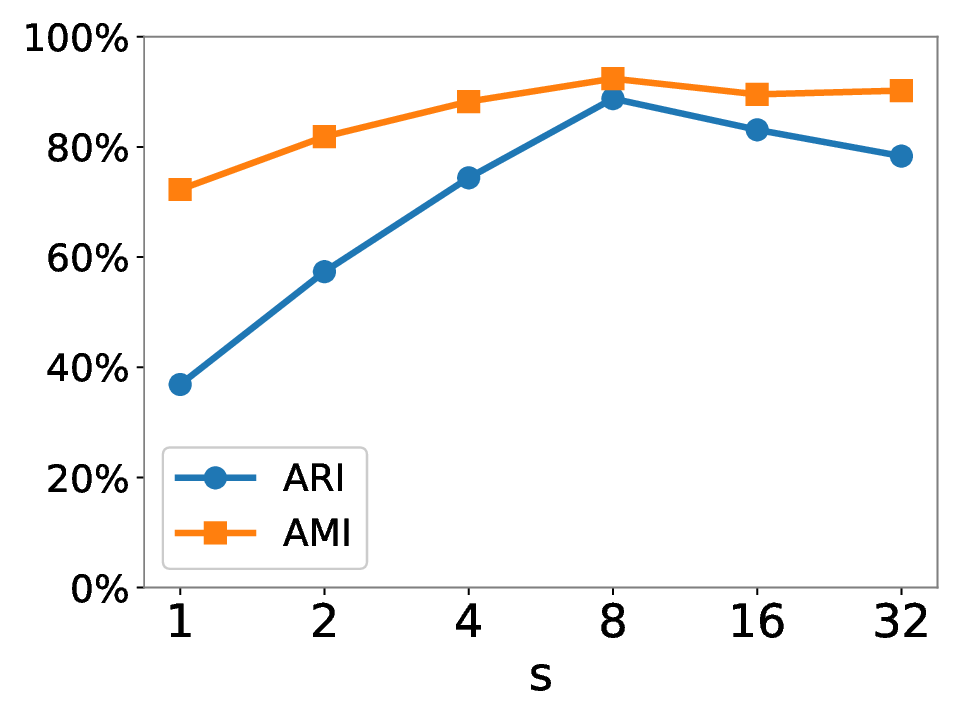}
    \end{minipage}
    \centering
    \caption{Parameter sensitivity of ASRC on UMIST. The results are presented in percentage format.}
    \label{para}
\end{figure}

\begin{itemize}
\item The parameter $\beta$ controls contrastive learning, and parameter $\lambda_2$ determines the graph structure learning. Both parameters are crucial for our method. An appropriate choice of $\beta$ and $\lambda_2$ enables the model to obtain improved representations, thereby boosting the clustering performance.

\item  The degree of sparsity $s$ and the number of iterations for updating graph structure $T_1$ play a critical role in determining the learned graph structure.
Values of $s$ that are either too small or too large can result in an inaccurate learned graph structure, failing to effectively reflect clustering information.
\end{itemize}

\subsubsection{Visualization of Clustering Results}
\begin{figure}[htbp]
    \centering
    \begin{minipage}[b]{0.15\textwidth}
        \centering
        \includegraphics[width=\textwidth]{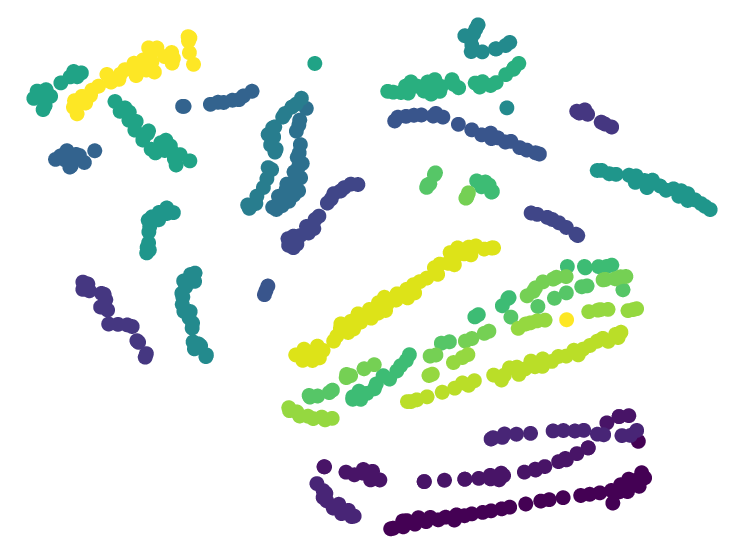}
        \caption*{(a) Raw features}
    \end{minipage}
    \begin{minipage}[b]{0.15\textwidth}
        \centering
        \includegraphics[width=\textwidth]{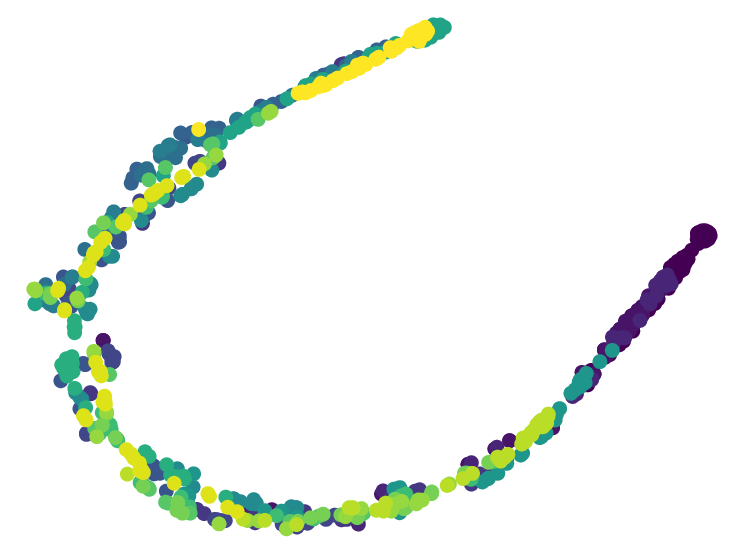}
        \caption*{(b) SDCN}
    \end{minipage}   
    \begin{minipage}[b]{0.15\textwidth}
        \centering
        \includegraphics[width=\textwidth]{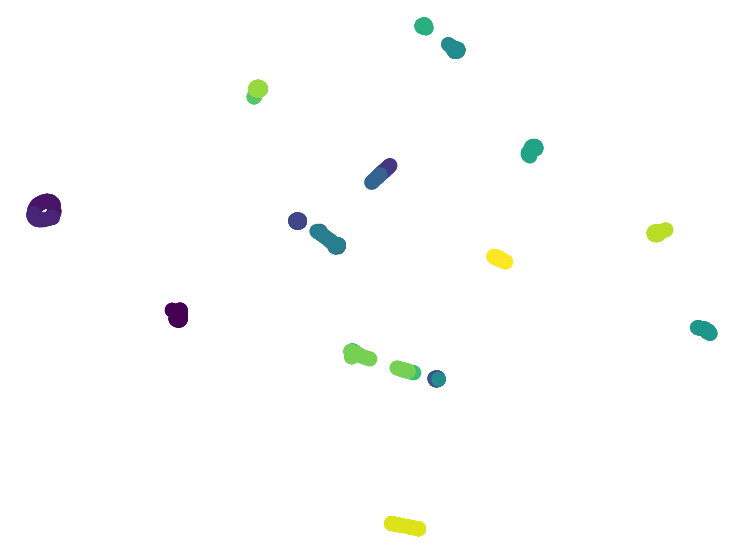}
        \caption*{(c) ASRC}
    \end{minipage}
    \centering
    \caption{t-SNE visualization results of the embedding on UMIST. (The true class number is $20$)}
    \label{visual}
\end{figure}

To deepen our comprehension of how embeddings influence clustering results, we use t-SNE to visualize the raw features and the embeddings obtained from SDCN and ASRC, as shown in Fig. \ref{visual}. Note that the embedding of ASRC is obtained from the GCN encoder, rather than the representatives from continuous clustering. We can observe that ASRC encourages instances of the same cluster (ground truth) to have similar embeddings, thereby achieving cohesive and separable clustering results.

\section{Related work}
\label{sec: related-work}

\subsection{Convex Clustering}
The convex clustering model and its variants have been extensively studied over the last ten years and have been widely applied in applications \cite{feng2023review}. On the other hand, the theoretical recovery guarantees of the convex clustering model have been established in both deterministic and statistical problem settings. The first recovery guarantee for the convex clustering was established in \cite{zhu2014convex} under relatively restrictive assumptions. Later, the deterministic recovery guarantees for the convex clustering model have been generalized to multi-cluster cases with more general geometry in \cite{panahi2017clustering}. However, these recovery guarantees are only for the convex clustering model with uniform weights. Note that Nguyen and Mamitsuka \cite{nguyen2021convex} proved that the convex clustering model with uniform weights can only recover convex clusters, where the convex hulls of each cluster are disjoint. Recently, the deterministic recovery guarantees for the general weighted convex clustering model have been established in \cite{sun2021convex}. From the statistical recovery guarantee perspective, the first statistical recovery guarantees of the convex clustering model with uniform weights were established in \cite{tan2015statistical}. The recovery guarantees have been generalized to general data distribution for the convex clustering model with $\ell_1$ norm fusion regularization in \cite{Radchenko2017} and to the Gaussian mixture data in \cite{jiang2020recovery}. More recently, some nice statistical guarantees for the weighted convex clustering model have been established by Dunlap and Mourrat \cite{dunlap2022local} for a more challenging star-shape geometry.

On the other hand, efficient optimization algorithms have been designed for solving large-scale convex cluster models. The alternating direction method of multipliers (ADMM) and the fast alternating minimization algorithm (AMA) are two popular first-order algorithms for solving the convex clustering model \cite{chi2015splitting}. However, the first-order algorithms are challenging to obtain solutions with high accuracy for the convex clustering model, especially for large-scale scenarios. To address this challenge, a semismooth Newton based augmented Lagrangian method (SSNAL) has been proposed in \cite{yuan2018efficient}. The fast convergence rate of the SSNAL and the unique ``second-order sparsity'' exploited by the semismooth Newont algorithm make this algorithm efficient for obtaining a highly accurate solution to the convex clustering model. Recently, dimension reduction techniques for the convex clustering model with a large number of data points \cite{yuan2021dimension} or high-dimensional data \cite{wang2023randomly} have been proposed to further accelerate solving this model.

Despite impressive progress on the convex clustering and the more general robust continuous cluster model, how to enhance the performance of these models in real applications with noisy unstructured data remains a great challenge, and this paper contributes to addressing this challenge by proposing a novel ASRC framework.

\subsection{Deep Clustering}

In recent years, deep clustering methods have attracted considerable attention due to their capacity to simultaneously learn representations and cluster data. The majority of deep clustering algorithms obtain embeddings by auto-encoders, followed by utilizing kmeans clustering on the learned embeddings to derive clustering outcomes. Unsupervised deep embedding for clustering analysis, as introduced in \cite{xie2016unsupervised}, leverages deep learning techniques to derive meaningful representations. Deep Embedded Clustering (DEC) \cite{xie2016unsupervised} utilizes the KL divergence loss to improve clustering cohesion. Improved Deep Embedded Clustering (IDEC) \cite{guo2017improved} has augmented DEC with reconstruction loss to facilitate better representation learning. Deep Clustering Network (DCN) \cite{yang2017towards} aims to learn representations aligned with clustering tasks. 

With the popularity of graph neural networks (GNNs) \cite{kipf2016semi}, in order to mine high-order structural information, clustering methods based on graph neural networks have also attracted widespread attention in recent years. The graph auto-encoder (GAE) \cite{kipf2016variational} exhibits robust encoding capabilities for graph data, enabling the acquisition of representations rich in information. Structural Deep Clustering Network (SDCN) \cite{bo2020structural} integrates auto-encoder and Graph Convolutional Network (GCN) \cite{kipf2016semi} to capture high-level information, offering an end-to-end clustering solution. Marginalized graph
auto-encoder (MGAE) \cite{wang2017mgae} leverages both structure and content information via the graph convolutional network (GCN) augmented auto-encoder, enhancing representation learning and achieving superior clustering performance. Embedding GAE (EGAE) \cite{zhang2022embedding} combines relaxed kmeans theory with graph auto-encoders, improving graph clustering and theoretical coherence. However, most of these methods are primarily focused on reconstructing adjacency matrices and are not applicable to unstructured data. Adaptive Graph Autoencoder (AdaGAE) \cite{li2021adaptive} learns the graph structure adaptively from unstructured data and combines it with GAE to encode data information. It aims to reconstruct the connectivity probability between nodes, achieving excellent performance on general data clustering tasks.

Additionally, there are some methods that combine contrastive learning to further improve the quality of feature representation. Specifically, Simple Contrastive
Graph Clustering (SCGC) \cite{liu2023simple} designs parameter-unshared siamese encoders and directly perturbs node embeddings to construct two augmented views of the same nodes for contrastive learning. Cluster-guided Contrastive deep Graph Clustering (CCGC) \cite{yang2023cluster} addresses limitations in existing contrastive learning methods by carefully constructing positive samples based on high-confidence clustering results.

However, they all rely on knowing the total number of clusters $c$ in advance and are sensitive to the initialization of cluster centers. When $c$ is not provided, none of the methods will work well.

\section{Conclusion}
\label{sec: conclusion}

In this paper, we propose a novel self-supervised deep clustering method, termed Adaptive Self-supervised Robust Clustering (ASRC), tailored for unstructured data without prior knowledge of the number of clusters. By adaptively learning the graph structure and utilizing the edge weights of the constructed graph, our method can effectively captures high-level structural information. Through the integration of GAE and contrastive learning, we refine the feature embedding to better align with the clustering objective. These enhancements strengthen the robust continuous clustering model by improving the graph structure, weight assignment, and data representation. Extensive experiments on benchmark datasets demonstrate the superior performance of our method, outperforming even those methods requiring prior knowledge of the number of clusters. Comprehensive ablation experiments further validate the effectiveness of our approach. In brief, ASRC offers a promising solution for clustering unstructured data with an unknown number of clusters.



\bibliographystyle{IEEEtran}
\bibliography{ASRC}

\begin{thebibliography}{10}
\providecommand{\url}[1]{#1}
\csname url@samestyle\endcsname
\providecommand{\newblock}{\relax}
\providecommand{\bibinfo}[2]{#2}
\providecommand{\BIBentrySTDinterwordspacing}{\spaceskip=0pt\relax}
\providecommand{\BIBentryALTinterwordstretchfactor}{4}
\providecommand{\BIBentryALTinterwordspacing}{\spaceskip=\fontdimen2\font plus
\BIBentryALTinterwordstretchfactor\fontdimen3\font minus \fontdimen4\font\relax}
\providecommand{\BIBforeignlanguage}[2]{{%
\expandafter\ifx\csname l@#1\endcsname\relax
\typeout{** WARNING: IEEEtran.bst: No hyphenation pattern has been}%
\typeout{** loaded for the language `#1'. Using the pattern for}%
\typeout{** the default language instead.}%
\else
\language=\csname l@#1\endcsname
\fi
#2}}
\providecommand{\BIBdecl}{\relax}
\BIBdecl

\bibitem{lloyd1982least}
S.~Lloyd, ``Least squares quantization in pcm,'' \emph{IEEE Transactions on Information Theory}, vol.~28, no.~2, pp. 129--137, 1982.

\bibitem{arthur2007k}
D.~Arthur and S.~Vassilvitskii, ``k-means++ the advantages of careful seeding,'' in \emph{Proceedings of the eighteenth annual ACM-SIAM symposium on Discrete algorithms}, 2007, pp. 1027--1035.

\bibitem{liu2010novel}
Q.~Liu, B.~Zhang, H.~Sun, Y.~Guan, and L.~Zhao, ``A novel k-means clustering algorithm based on positive examples and careful seeding,'' in \emph{2010 International Conference on Computational and Information Sciences}.\hskip 1em plus 0.5em minus 0.4em\relax IEEE, 2010, pp. 767--770.

\bibitem{hamalainen2020improving}
J.~H{\"a}m{\"a}l{\"a}inen, T.~K{\"a}rkk{\"a}inen, and T.~Rossi, ``Improving scalable k-means++,'' \emph{Algorithms}, vol.~14, no.~1, p.~6, 2020.

\bibitem{celebi2013comparative}
M.~E. Celebi, H.~A. Kingravi, and P.~A. Vela, ``A comparative study of efficient initialization methods for the k-means clustering algorithm,'' \emph{Expert systems with applications}, vol.~40, no.~1, pp. 200--210, 2013.

\bibitem{ng2001spectral}
A.~Ng, M.~Jordan, and Y.~Weiss, ``On spectral clustering: Analysis and an algorithm,'' \emph{Advances in Neural Information Processing Systems}, vol.~14, 2001.

\bibitem{von2008consistency}
U.~Von~Luxburg, M.~Belkin, and O.~Bousquet, ``Consistency of spectral clustering,'' \emph{The Annals of Statistics}, pp. 555--586, 2008.

\bibitem{lei2015consistency}
J.~Lei and A.~Rinaldo, ``Consistency of spectral clustering in stochastic block models,'' 2015.

\bibitem{von2007tutorial}
U.~Von~Luxburg, ``A tutorial on spectral clustering,'' \emph{Statistics and computing}, vol.~17, pp. 395--416, 2007.

\bibitem{rodriguez2014clustering}
A.~Rodriguez and A.~Laio, ``Clustering by fast search and find of density peaks,'' \emph{science}, vol. 344, no. 6191, pp. 1492--1496, 2014.

\bibitem{campello2013density}
R.~J. Campello, D.~Moulavi, and J.~Sander, ``Density-based clustering based on hierarchical density estimates,'' in \emph{Pacific-Asia conference on knowledge discovery and data mining}.\hskip 1em plus 0.5em minus 0.4em\relax Springer, 2013, pp. 160--172.

\bibitem{ester1996density}
M.~Ester, H.-P. Kriegel, J.~Sander, X.~Xu \emph{et~al.}, ``A density-based algorithm for discovering clusters in large spatial databases with noise,'' in \emph{kdd}, vol.~96, no.~34, 1996, pp. 226--231.

\bibitem{johnson1967hierarchical}
S.~C. Johnson, ``Hierarchical clustering schemes,'' \emph{Psychometrika}, vol.~32, no.~3, pp. 241--254, 1967.

\bibitem{cheng2019hierarchical}
D.~Cheng, Q.~Zhu, J.~Huang, Q.~Wu, and L.~Yang, ``A hierarchical clustering algorithm based on noise removal,'' \emph{International Journal of Machine Learning and Cybernetics}, vol.~10, pp. 1591--1602, 2019.

\bibitem{rokach2005clustering}
L.~Rokach and O.~Maimon, ``Clustering methods,'' \emph{Data mining and knowledge discovery handbook}, pp. 321--352, 2005.

\bibitem{liu2023reinforcement}
Y.~Liu, K.~Liang, J.~Xia, X.~Yang, S.~Zhou, M.~Liu, X.~Liu, and S.~Z. Li, ``Reinforcement graph clustering with unknown cluster number,'' in \emph{Proceedings of the 31st ACM International Conference on Multimedia}, 2023, pp. 3528--3537.

\bibitem{pelckmans2005convex}
K.~Pelckmans, J.~De~Brabanter, J.~A. Suykens, and B.~De~Moor, ``Convex clustering shrinkage,'' in \emph{PASCAL Workshop on Statistics and Optimization of Clustering Workshop}, 2005.

\bibitem{hocking2011clusterpath}
T.~D. Hocking, A.~Joulin, F.~Bach, and J.-P. Vert, ``Clusterpath an algorithm for clustering using convex fusion penalties,'' in \emph{28th International Conference on Machine Learning}, 2011, pp. 745--752.

\bibitem{lindsten2011clustering}
F.~Lindsten, H.~Ohlsson, and L.~Ljung, ``Clustering using sum-of-norms regularization: With application to particle filter output computation,'' in \emph{2011 IEEE Statistical Signal Processing Workshop}, 2011, pp. 201--204.

\bibitem{chi2015splitting}
E.~C. Chi and K.~Lange, ``Splitting methods for convex clustering,'' \emph{Journal of Computational and Graphical Statistics}, vol.~24, no.~4, pp. 994--1013, 2015.

\bibitem{shah2017robust}
S.~A. Shah and V.~Koltun, ``Robust continuous clustering,'' \emph{Proceedings of the National Academy of Sciences}, vol. 114, no.~37, pp. 9814--9819, 2017.

\bibitem{shah2018deep}
------, ``Deep continuous clustering,'' \emph{arXiv preprint arXiv:1803.01449}, 2018.

\bibitem{sun2021convex}
D.~Sun, K.-C. Toh, and Y.~Yuan, ``Convex clustering: Model, theoretical guarantee and efficient algorithm,'' \emph{The Journal of Machine Learning Research}, vol.~22, no.~1, pp. 427--458, 2021.

\bibitem{nguyen2021convex}
C.~H. Nguyen and H.~Mamitsuka, ``On convex clustering solutions,'' \emph{arXiv preprint arXiv:2105.08348}, 2021.

\bibitem{bo2020structural}
D.~Bo, X.~Wang, C.~Shi, M.~Zhu, E.~Lu, and P.~Cui, ``Structural deep clustering network,'' in \emph{Proceedings of the web conference 2020}, 2020, pp. 1400--1410.

\bibitem{dunlap2022local}
A.~Dunlap and J.-C. Mourrat, ``Local versions of sum-of-norms clustering,'' \emph{SIAM Journal on Mathematics of Data Science}, vol.~4, no.~4, pp. 1250--1271, 2022.

\bibitem{geman1987statistical}
S.~Geman, ``Statistical methods for tomographic image restoration,'' \emph{Bull. Internat. Statist. Inst.}, vol.~52, pp. 5--21, 1987.

\bibitem{li2021adaptive}
X.~Li, H.~Zhang, and R.~Zhang, ``Adaptive graph auto-encoder for general data clustering,'' \emph{IEEE Transactions on Pattern Analysis and Machine Intelligence}, vol.~44, no.~12, pp. 9725--9732, 2021.

\bibitem{wu2021self}
J.~Wu, X.~Wang, F.~Feng, X.~He, L.~Chen, J.~Lian, and X.~Xie, ``Self-supervised graph learning for recommendation,'' in \emph{Proceedings of the 44th international ACM SIGIR conference on research and development in information retrieval}, 2021, pp. 726--735.

\bibitem{nie2014clustering}
F.~Nie, X.~Wang, and H.~Huang, ``Clustering and projected clustering with adaptive neighbors,'' in \emph{Proceedings of the 20th ACM SIGKDD international conference on Knowledge discovery and data mining}, 2014, pp. 977--986.

\bibitem{panahi2017clustering}
A.~Panahi, D.~Dubhashi, F.~D. Johansson, and C.~Bhattacharyya, ``Clustering by sum of norms: Stochastic incremental algorithm, convergence and cluster recovery,'' in \emph{34th International Conference on Machine Learning}, 2017, pp. 2769--2777.

\bibitem{zhao2021graph}
H.~Zhao, X.~Yang, Z.~Wang, E.~Yang, and C.~Deng, ``Graph debiased contrastive learning with joint representation clustering.'' in \emph{IJCAI}, 2021, pp. 3434--3440.

\bibitem{liu2023hard}
Y.~Liu, X.~Yang, S.~Zhou, X.~Liu, Z.~Wang, K.~Liang, W.~Tu, L.~Li, J.~Duan, and C.~Chen, ``Hard sample aware network for contrastive deep graph clustering,'' in \emph{Proceedings of the AAAI conference on artificial intelligence}, vol.~37, no.~7, 2023, pp. 8914--8922.

\bibitem{hou2013joint}
C.~Hou, F.~Nie, X.~Li, D.~Yi, and Y.~Wu, ``Joint embedding learning and sparse regression: A framework for unsupervised feature selection,'' \emph{IEEE transactions on cybernetics}, vol.~44, no.~6, pp. 793--804, 2013.

\bibitem{nene1996columbia}
S.~A. Nene, S.~K. Nayar, H.~Murase \emph{et~al.}, ``Columbia object image library (coil-20),'' 1996.

\bibitem{lyons1999automatic}
M.~J. Lyons, J.~Budynek, and S.~Akamatsu, ``Automatic classification of single facial images,'' \emph{IEEE transactions on pattern analysis and machine intelligence}, vol.~21, no.~12, pp. 1357--1362, 1999.

\bibitem{asuncion2007uci}
A.~Asuncion and D.~Newman, ``Uci machine learning repository,'' 2007.

\bibitem{hartigan1979algorithm}
J.~A. Hartigan and M.~A. Wong, ``Algorithm as 136: A k-means clustering algorithm,'' \emph{Journal of the royal statistical society. series c (applied statistics)}, vol.~28, no.~1, pp. 100--108, 1979.

\bibitem{shi2000normalized}
J.~Shi and J.~Malik, ``Normalized cuts and image segmentation,'' \emph{IEEE Transactions on pattern analysis and machine intelligence}, vol.~22, no.~8, pp. 888--905, 2000.

\bibitem{shaham2018spectralnet}
U.~Shaham, K.~Stanton, H.~Li, B.~Nadler, R.~Basri, and Y.~Kluger, ``Spectralnet: Spectral clustering using deep neural networks,'' \emph{arXiv preprint arXiv:1801.01587}, 2018.

\bibitem{kipf2016variational}
T.~N. Kipf and M.~Welling, ``Variational graph auto-encoders,'' \emph{arXiv preprint arXiv:1611.07308}, 2016.

\bibitem{li2022maskgae}
J.~Li, R.~Wu, W.~Sun, L.~Chen, S.~Tian, L.~Zhu, C.~Meng, Z.~Zheng, and W.~Wang, ``Maskgae: Masked graph modeling meets graph autoencoders,'' \emph{arXiv preprint arXiv:2205.10053}, vol.~9, p.~13, 2022.

\bibitem{feng2023review}
Q.~Feng, C.~P. Chen, and L.~Liu, ``A review of convex clustering from multiple perspectives: models, optimizations, statistical properties, applications, and connections,'' \emph{IEEE Transactions on Neural Networks and Learning Systems}, 2023.

\bibitem{zhu2014convex}
C.~Zhu, H.~Xu, C.~Leng, and S.~Yan, ``Convex optimization procedure for clustering: Theoretical revisit,'' \emph{Advances in Neural Information Processing Systems}, vol.~27, pp. 1619--1627, 2014.

\bibitem{tan2015statistical}
K.~M. Tan and D.~Witten, ``Statistical properties of convex clustering,'' \emph{Electronic Journal of Statistics}, vol.~9, no.~2, pp. 2324--2347, 2015.

\bibitem{Radchenko2017}
P.~Radchenko and G.~Mukherjee, ``Convex clustering via {$l_1$} fusion penalization,'' \emph{Journal of the Royal Statistical Society. Series B (Statistical Methodology)}, vol.~79, no.~5, pp. 1527--1546, 2017.

\bibitem{jiang2020recovery}
T.~Jiang, S.~Vavasis, and C.~W. Zhai, ``Recovery of a mixture of {G}aussians by sum-of-norms clustering,'' \emph{Journal of Machine Learning Research}, vol.~21, no. 225, pp. 1--16, 2020.

\bibitem{yuan2018efficient}
Y.~Yuan, D.~Sun, and K.-C. Toh, ``An efficient semismooth {N}ewton based algorithm for convex clustering,'' in \emph{35th International Conference on Machine Learning}, 2018, pp. 5718--5726.

\bibitem{yuan2021dimension}
Y.~Yuan, T.-H. Chang, D.~Sun, and K.-C. Toh, ``A dimension reduction technique for large-scale structured sparse optimization problems with application to convex clustering,'' \emph{SIAM Journal on Optimization}, vol.~32, no.~3, pp. 2294--2318, 2022.

\bibitem{wang2023randomly}
Z.~Wang, Y.~Yuan, J.~Ma, T.~Zeng, and D.~Sun, ``Randomly projected convex clustering model: Motivation, realization, and cluster recovery guarantees,'' \emph{arXiv preprint arXiv:2303.16841}, 2023.

\bibitem{xie2016unsupervised}
J.~Xie, R.~Girshick, and A.~Farhadi, ``Unsupervised deep embedding for clustering analysis,'' in \emph{International conference on machine learning}.\hskip 1em plus 0.5em minus 0.4em\relax PMLR, 2016, pp. 478--487.

\bibitem{guo2017improved}
X.~Guo, L.~Gao, X.~Liu, and J.~Yin, ``Improved deep embedded clustering with local structure preservation.'' in \emph{Ijcai}, vol.~17, 2017, pp. 1753--1759.

\bibitem{yang2017towards}
B.~Yang, X.~Fu, N.~D. Sidiropoulos, and M.~Hong, ``Towards k-means-friendly spaces: Simultaneous deep learning and clustering,'' in \emph{international conference on machine learning}.\hskip 1em plus 0.5em minus 0.4em\relax PMLR, 2017, pp. 3861--3870.

\bibitem{kipf2016semi}
T.~N. Kipf and M.~Welling, ``Semi-supervised classification with graph convolutional networks,'' \emph{arXiv preprint arXiv:1609.02907}, 2016.

\bibitem{wang2017mgae}
C.~Wang, S.~Pan, G.~Long, X.~Zhu, and J.~Jiang, ``Mgae: Marginalized graph autoencoder for graph clustering,'' in \emph{Proceedings of the 2017 ACM on Conference on Information and Knowledge Management}, 2017, pp. 889--898.

\bibitem{zhang2022embedding}
H.~Zhang, P.~Li, R.~Zhang, and X.~Li, ``Embedding graph auto-encoder for graph clustering,'' \emph{IEEE Transactions on Neural Networks and Learning Systems}, 2022.

\bibitem{liu2023simple}
Y.~Liu, X.~Yang, S.~Zhou, X.~Liu, S.~Wang, K.~Liang, W.~Tu, and L.~Li, ``Simple contrastive graph clustering,'' \emph{IEEE Transactions on Neural Networks and Learning Systems}, 2023.

\bibitem{yang2023cluster}
X.~Yang, Y.~Liu, S.~Zhou, S.~Wang, W.~Tu, Q.~Zheng, X.~Liu, L.~Fang, and E.~Zhu, ``Cluster-guided contrastive graph clustering network,'' in \emph{Proceedings of the AAAI conference on artificial intelligence}, vol.~37, no.~9, 2023, pp. 10\,834--10\,842.

\end{thebibliography}

\end{document}


\title{Appendix for Adaptive Self-supervised Robust Clustering for Unstructured Data with Unknown Cluster Number}

\author{Chen-Lu Ding\textsuperscript{$*$},
        Jiancan Wu\textsuperscript{$*$},
        Wei Lin,
        Shiyang Shen,
        Xiang Wang\textsuperscript{$\dagger$},
        Yancheng Yuan\textsuperscript{$\dagger$}
\thanks{$*$: Chen-Lu Ding and Jiancan Wu contribute equally to this manuscript.}
\thanks{$\dagger$: Xiang Wang and Yancheng Yuan are Corresponding Authors.}
\thanks{Chen-Lu Ding, Jiancan Wu, Wei Lin, and Xiang Wang are with University of Science and Technology of China. E-mail: dingchenlu200103@gmail.com, wujcan@gmail.com, kkwml@mail.ustc.edu.cn, xiangwang@ustc.edu.cn.}
\thanks{Shiyang Shen and Yancheng Yuan are with The Hong Kong Polytechnic University. E-mail: 22049485g@connect.polyu.hk, yancheng.yuan@polyu.edu.hk.}
}

\markboth{Journal of \LaTeX\ Class Files,~Vol.~0000, No.~0000}%
{Shell \MakeLowercase{\textit{et al.}}: A Sample Article Using IEEEtran.cls for IEEE Journals}

\maketitle



{\appendix 

\section*{Appendix A: Complete ablation experiment results.}
\label{comapp}
In this section, we provide our complete ablation experiment results, which can be found in Table \ref{complete}. We can observe that ASRC consistently achieves the best performance across all seven datasets. In summary, by integrating improved adaptive graph structure learning and unbiased negative sampling, ASRC achieves superior clustering results through the acquisition of robust graph structures, consistent weight distributions, and discriminative data representations.
\begin{table*}[htbp]
\renewcommand{\arraystretch}{1.3}
\caption[]{Clustering results with different variants on seven datasets (mean±std). The best results in each case are marked in bold. The results are presented in percentage format.}
\label{complete}
\begin{center}
\begin{tabular}{@{\hspace{1.5em}}c@{\hspace{1em}}|@{\hspace{2em}}c@{\hspace{2em}}|@{\hspace{4em}}c@{\hspace{4em}}c@{\hspace{4em}}c@{\hspace{4em}}c@{\hspace{4em}}c@{\hspace{3em}}}
\specialrule{1pt}{0pt}{0pt}
Datasets     & Metric & RCC        & AdaGAE       &   ASRC-1     &  ASRC-2       & ASRC             \\ \hline
\multirow{2}{*}{20NEWS}               & AMI    & 0.03±0.00  &  15.22±0.00  &  29.32±0.01  &   30.06±0.01 &  \textbf{32.10±0.00} \\
             & ARI    & 0.27±0.00  &  13.83±0.00  &  23.83±0.02  &   24.41±0.02 &  \textbf{26.55±0.00}\\ \hline
             
\multirow{2}{*}{UMIST}               & AMI    & 79.47±0.00 &  90.26±0.01  &  91.28±0.00  &   91.42±0.00 &  \textbf{94.42±0.00} \\
             & ARI    & 52.32±0.00 &  79.52±0.02  &  79.69±0.01  &   82.22±0.01 &  \textbf{89.65±0.01}\\ \hline
             
\multirow{2}{*}{Mice Protein}               & AMI    & 58.24±0.00 &  56.83±0.01  &  65.28±0.02  &   68.66±0.01 &  \textbf{70.20±0.02}\\
             & ARI    & 21.50±0.00 &  37.84±0.02  &  37.34±0.01  &   44.37±0.01 &  \textbf{45.39±0.02}\\ \hline

\multirow{2}{*}{COIL-20}               & AMI    & 84.24±0.00 &  97.76±0.00  &  96.98±0.00  &   97.10±0.00 &  \textbf{97.97±0.00}\\
             & ARI    & 76.36±0.00 &  94.75±0.00  &  94.64±0.00  &   94.93±0.00 &  \textbf{96.24±0.00}\\ \hline

\multirow{2}{*}{Jaffe}               & AMI    & 92.21±0.00 &  94.44±0.00  &  95.01±0.00  &   96.17±0.00 &  \textbf{96.39±0.00}\\
             & ARI    & 87.70±0.00 &  91.50±0.00  &  92.27±0.00  &   95.35±0.01 &  \textbf{95.57±0.00}\\ \hline

\multirow{2}{*}{USPS}               & AMI    & 79.55±0.00 &  82.94±0.01  &  83.78±0.01  &   82.93±0.00 &  \textbf{84.12±0.00}\\
             & ARI    & 73.21±0.00 &  76.33±0.01  &  78.67±0.02  &   76.24±0.01 &  \textbf{79.56±0.00}\\ \hline

\multirow{2}{*}{MNIST}               & AMI    & 78.93±0.00 &  81.90±0.01  &  82.16±0.01  &   82.16±0.01 &  \textbf{82.54±0.00}\\
             & ARI    & 60.54±0.00 &  74.59±0.03  &  77.02±0.02  &   77.02±0.01 &  \textbf{77.19±0.01}\\ \hline
             
\specialrule{1pt}{0pt}{0pt}
\end{tabular}
\end{center}
\end{table*}

\section*{Appendix B: The parameter settings of ASRC.}
\label{paraapp}
We provide the optimal parameters for ASRC in Table \ref{setting}, where the meanings of each parameter are given as follows: $k_0$ denotes initial sparsity; $s$ represents the increment of sparsity; $T_1$ indicates the number of iterations for updating the graph structure; $\beta$ is the trade-off parameter in Eq.(16); struct refers to the network structure of GCN; and $T_2$ specifies the number of iterations for updating the graph structure with fixed $k$ values.

\begin{table}[htbp]
\caption[]{The parameter settings.}
\label{setting}
\renewcommand{\arraystretch}{1.1}
\small 

\begin{tabular}{cp{0.2em}ccp{0.2em}ccccp{0.4em}c} 
\toprule
\textbf{Dataset}      & \textbf{$k_0$}  & \textbf{$s$}  & \textbf{$\lambda_2$}  & \textbf{$T_1$} & \textbf{$\beta$}   & {struct}   & \textbf{$T_2$} \\
\midrule
\textbf{20NEWS}       & 50 & 150 & $2^{-3}$   & 7  & $10^{-3}$  & $d$-256-64 & $2$\\
\textbf{UMIST}        & 5  & 8   & 4     & 7  & 10     & $d$-256-64 & $2$\\
\textbf{COIL-20}      & 5  & 8   & 8     & 10 & 1      & $d$-256-64 & $1$\\
\textbf{MNIST}        & 10  & 64  & $2^{-6}$ & 15 & $10^{-3}$ & $d$-256-64 & $2$\\
\textbf{JAFFE}        & 15  & 2   & $2^{-6}$ & 10 & 1      & $d$-256-64 & $4$\\
\textbf{Mice Protein} & 10  & 2   & $2^{-6}$ & 20 & 1      & $d$-256-64 & $2$\\
\textbf{USPS}         & 10  & 50  & 4     & 7 & 10     & d-128-64 & $2$\\ \bottomrule
\end{tabular}
\end{table}

\section*{Appendix C: Analysis of PCA n components.}
\label{pcaapp}
\begin{figure}[htbp]
\centering
\includegraphics[width=0.4\textwidth]{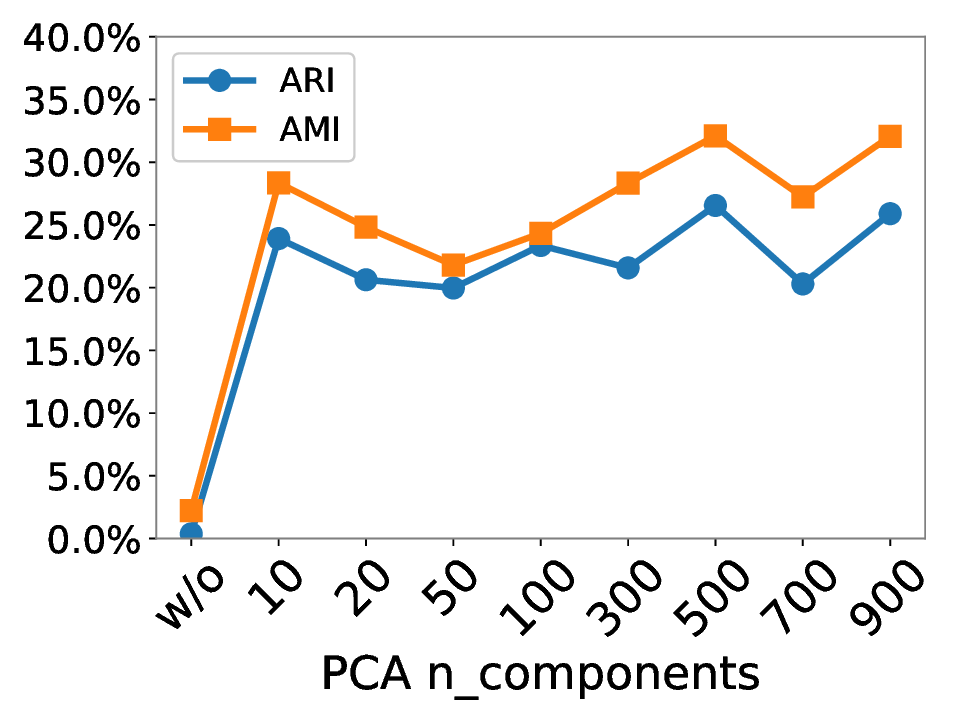}
\caption{Clustering results with different PCA components on 20NEWS. w/o refers to without PCA preprocessing. The results are presented in percentage format.}
\label{pcafi}
\end{figure}

The clutserig results of different PCA components on 20NEWS dataset is shown in Figure \ref{pcafi}. Without PCA preprocessing, ASRC takes raw text features and cannot learn informative and discriminative representations. Since continuous clustering relies on the distance of representatives to control the cluster number, it is more sensitive to distance, and the component of PCA has a certain impact on the clustering results. For 20NEWS, we selected the components of PCA to be $500$.

\section*{Appendix D: A Solution to the Probability Learning with General Priors and Its Implications.}
\label{solutionapp}
In this section, we give a solution to the following nonconvex optimization problem
\begin{equation}
\label{app_pl}
\min_{\boldsymbol{p} \in \Delta_n, \|\boldsymbol{p}\|_0 \leq k} ~ \left\langle\boldsymbol{p}, \boldsymbol{d}\right\rangle+\frac{\gamma}{2}\left\|\boldsymbol{p} - \boldsymbol{q}\right\|_2^2,
\end{equation}
where $\boldsymbol{d} \in \mathbb{R}^n$ is a given vector, $\boldsymbol{q} \in \Delta_n$ is a given prior distribution, $\gamma > 0$ and the integer $k$ are given hyper-parameters. Next, we will give a solution to the nonconvex optimization problem \eqref{app_pl}. As a direct consequence, this will also give a solution to (8) due to its separable structure. 

For convenience, we introduce some notation. Let $\boldsymbol{a} \in \mathbb{R}^n$ be any given vector. Let $I \subseteq \{1, 2, \dots, n\}$ be a given index set. Denote the subvector of $\boldsymbol{a}$ with elements indexed by $I$ as $\boldsymbol{a}_I$. Denote $\boldsymbol{a}^{\downarrow} := (\boldsymbol{a}_{i_1}, \dots, \boldsymbol{a}_{i_n})$ as the vector whose elements are rearranging $\boldsymbol{a}$ in descending order and $\mathcal{I}_k^{\downarrow}(\boldsymbol{a}) := \{i_1, \dots, i_k\}$. Denote 
\[
(\mathcal{P}_{k}(\boldsymbol{a}))_i := \left\{
\begin{array}{ll}
 \boldsymbol{a}_i    &  \mbox{if $i \in \mathcal{I}^{\downarrow}_k(\boldsymbol{a})$},\\
 0    & \mbox{otherwise}. 
\end{array}
\right.
\]
Denote $\mbox{supp}(\boldsymbol{a}) := \{1 \leq i \leq n ~|~ \boldsymbol{a}_i \neq 0\}$.

It follows from \cite[Theorem 1]{pmlr-v28-kyrillidis13} that the solution $\boldsymbol{p}^*$ to \eqref{app_pl} can be found via Algorithm \ref{algo1-pl}, where the projection onto the simplex set can be found in Algorithm \ref{algo2}.

\begin{algorithm}[htbp]
\caption{Solution to \eqref{app_pl}}\label{algo1-pl}
\begin{algorithmic}[1]
\STATE \textbf{Input}: $\boldsymbol{d} \in \mathbb{R}^n$, $\boldsymbol{q} \in \mathbb{R}^n$, $\gamma > 0$, $1 \leq k \leq n$.
\STATE \textbf{Output}: $\boldsymbol{p}^* \in \mathbb{R}^n$.
\STATE \textbf{Step 1}: Compute 
\[
\boldsymbol{c} = \boldsymbol{q} - \boldsymbol{d}/\gamma.
\]
\STATE \textbf{Step 2}: Sort the vector $\boldsymbol{c}$ in descending order to obtain $\boldsymbol{c}^{\downarrow}$ and
\[
\mathcal{S}^* := \mathcal{I}_k^{\downarrow}(\boldsymbol{c}), \quad (\mathcal{S}^*)^c := \{1, \dots, n\} \backslash \mathcal{S}^*. 
\]
\STATE \textbf{Step 3}: Compute 
\[
\boldsymbol{p}^*_{\mathcal{S}^*} = \Pi_{\Delta_{k}}(\boldsymbol{c}_{\mathcal{S}^*}), \quad \boldsymbol{p}^*_{(\mathcal{S}^*)^c} = 0.
\]
\STATE \textbf{Return}: $\boldsymbol{p}^*$.
\end{algorithmic}
\end{algorithm}

\begin{algorithm}[htbp]
\caption{Compute $\Pi_{\Delta_k}(\boldsymbol{c})$}\label{algo2}
\begin{algorithmic}[1]
\STATE \textbf{Input}: $\boldsymbol{c} \in \mathbb{R}^k$.
\STATE \textbf{Output}: $\boldsymbol{c}^* \in \mathbb{R}^k$.
\STATE \textbf{Step 1}: Sort $\boldsymbol{c}$ in descending order 
\[
\boldsymbol{c}^{\downarrow}_1 \geq \boldsymbol{c}^{\downarrow}_2 \geq \cdots \boldsymbol{c}^{\downarrow}_k .
\]
\STATE \textbf{Step 2}: Find the index 
\[
r = \max\left\{1 \leq j \leq k ~|~ \boldsymbol{c}^{\downarrow}_j - \frac{1}{j}\left(\sum_{i = 1}^j \boldsymbol{c}^{\downarrow}_i - 1\right) > 0 \right\}.
\]
\STATE \textbf{Step 3}: Define
\[
\theta = \frac{1}{r}\left(\sum_{i = 1}^r \boldsymbol{c}^{\downarrow}_i - 1\right).
\]
\STATE \textbf{Step 4}: Compute $\boldsymbol{c}^*$ by 
\[
\boldsymbol{c}^*_i = \max\{\boldsymbol{c}_i - \theta, 0\}, \quad 1 \leq i \leq k.
\]
\STATE \textbf{Return}: $\boldsymbol{c}^*$.
\end{algorithmic}
\end{algorithm}

Indeed, the solution constructed in Algorithm \ref{algo1-pl} to \eqref{app_pl} coincides with (10) if $\boldsymbol{q}$ is the uniform distribution. In applications, we prefer to keep the homogeneity of the graph, we require 
\begin{equation}
\label{eq: homogeneity}
\boldsymbol{p}^*_i \geq \boldsymbol{p}^*_j  \quad \mbox{if} \quad \boldsymbol{d}_i \leq \boldsymbol{d}_j.    
\end{equation}
Therefore, we want to choose appropriate prior distribution $\boldsymbol{q} \in \Delta_n$ and $\gamma > 0$, such that \eqref{eq: homogeneity} holds. It follows from Algorithm \ref{algo1-pl} and Algorithm \ref{algo2} that, we requires
\[
\boldsymbol{c}_i \geq \boldsymbol{c}_j.
\]
By some simple calculation, we know 
\[
\boldsymbol{q}_i - \boldsymbol{q}_j \geq (\boldsymbol{d}_i - \boldsymbol{d}_j)/\gamma.
\]
Therefore, the homogeneity of the graph can be preserved for any $\gamma > 0$ if $\boldsymbol{q}$ is chosen to be a uniform distribution.
}

\bibliographystyle{IEEEtran}
\bibliography{ASRC}